\documentclass[10pt,twocolumn,letterpaper]{article}

\usepackage{cvpr}
\usepackage{times}
\usepackage{epsfig}
\usepackage{graphicx}
\usepackage{amsmath}
\usepackage{amssymb}
\usepackage[nice]{nicefrac}

\usepackage{caption}
\usepackage{amsfonts}
\usepackage{bbm}

\usepackage{subcaption}
\usepackage{color,soul}
\usepackage{multirow}
\usepackage{floatrow}
\usepackage{float}
\usepackage{makecell}
\definecolor{shadecolor}{rgb}{0.92,0.92,0.92}
\usepackage{framed}
\usepackage{cite,xspace}

\newcommand{\code}[1]{{\ensuremath{\tt #1}}} 

\def\OurMethod{{BlendMask}\xspace}

\usepackage[pagebackref=true,breaklinks=true,colorlinks,bookmarks=false]{hyperref}
\cvprfinalcopy %

\def\vs{{\it vs.}}

\begin{document}

	\title{\OurMethod:
		Top-Down Meets Bottom-Up for Instance Segmentation%
\thanks{ indicates equal contributions.
 		 K. Sun's contribution  was made when
 		 visiting The University of Adelaide.
 	}
	}

	\author{
	Hao Chen$ ^{1*} $,  ~
	Kunyang Sun$ ^{2,1*} $, ~
		Zhi Tian$ ^1 $,  ~
		Chunhua Shen$ ^1 $\thanks{E-mail: $\tt  chunhua.shen@adelaide.edu.au $},  ~
		Yongming Huang$ ^2 $, ~
		Youliang Yan$ ^3 $
		\\[.125cm]
		$ ^1 $ The University of Adelaide, Australia
		~~
		$ ^2 $ Southeast University, China
		~~
		$ ^3 $ Huawei Noah's Ark Lab
	}

\maketitle
	
\begin{abstract}
Instance segmentation is one of the fundamental
vision tasks.
Recently, fully convolutional instance segmentation methods
have drawn much attention as they are often
simpler and more efficient than two-stage approaches like Mask R-CNN.
To date,
almost all such approaches
 fall behind the two-stage
Mask R-CNN method
in mask precision
when models have similar computation complexity,
leaving great room for improvement.
In this work,  we
 achieve improved mask prediction by  effectively combining
instance-level information with %
semantic information
with lower-level fine-granularity.
Our main contribution is a blender module which draws inspiration from both top-down and bottom-up instance segmentation approaches. The proposed \OurMethod can effectively predict dense per-pixel position-sensitive instance features with very few channels, and learn
attention maps for each instance
with merely one convolution layer, thus being fast in inference.
\OurMethod
can be easily incorporated  with
the state-of-the-art one-stage detection frameworks  and outperforms Mask R-CNN under the same training schedule while being 20\% faster.
A light-weight
version of \OurMethod
achieves $ 34.2\% $  mAP at 25 FPS evaluated on a single 1080Ti GPU card. Because of its simplicity and efficacy, we hope that our \OurMethod could serve as a simple yet strong baseline for a wide range of
instance-wise prediction
tasks.

Code is available at https://git.io/AdelaiDet
\end{abstract}

\section{Introduction}

The top performing object detectors and segmenters often follow a two-stage paradigm. They consist of a fully convolutional network,
region proposal network (RPN), to perform dense prediction of the most likely regions of interest (RoIs). A set of light-weight networks,
a.k.a.\
heads, are applied to
re-align
the features of RoIs and generate predictions \cite{ren2015faster}. The quality and speed for mask generation is strongly tied
to
the structure of the mask heads. In addition, it is difficult for independent heads to share features with related tasks such as semantic segmentation which causes trouble for network architecture optimization.

\begin{figure}[t]
\centering
\includegraphics[width=\textwidth]{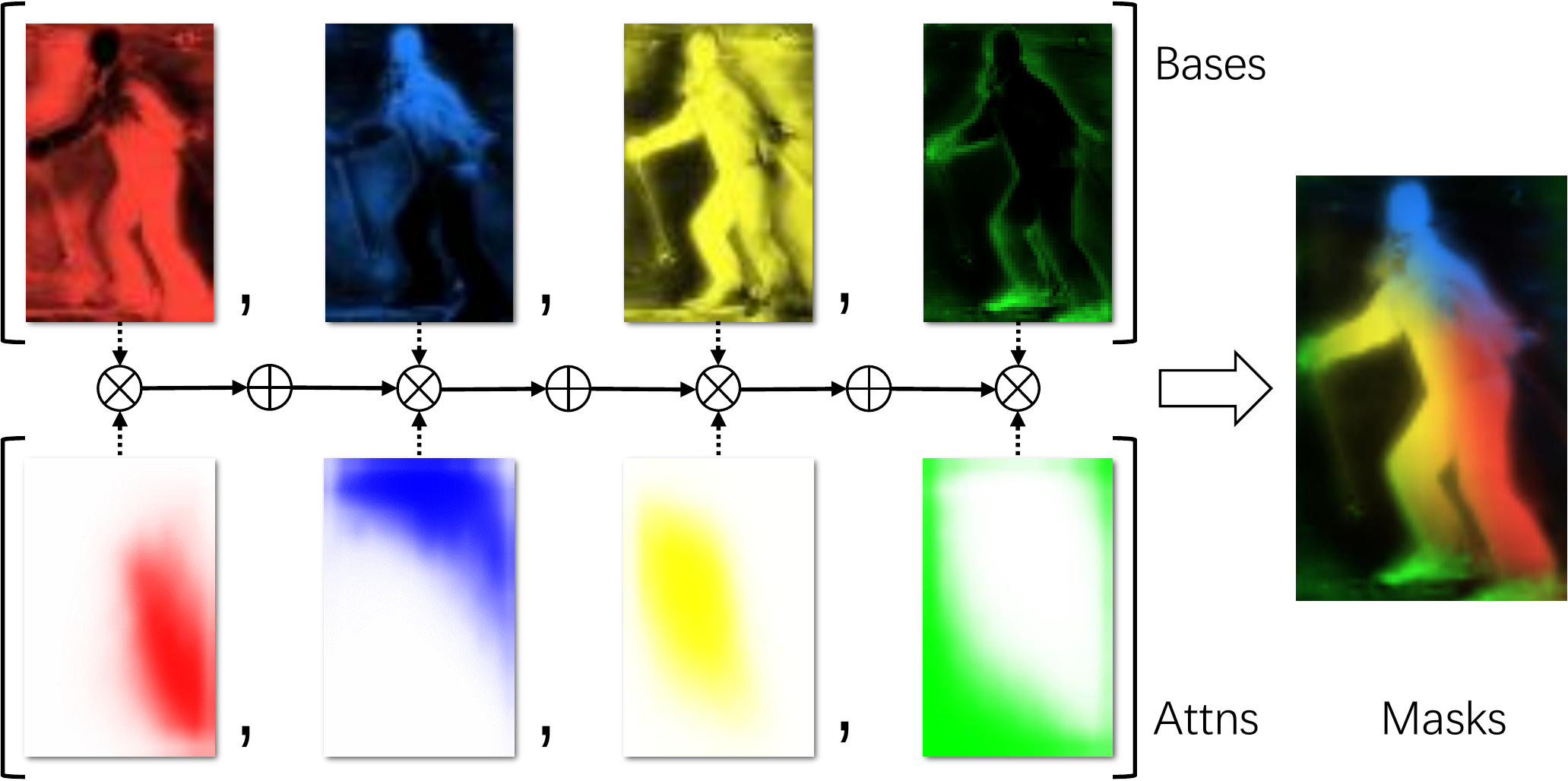}
\caption{\textbf{Blending process}. We illustrate an example of the learned bases and attentions.
Four bases and attention maps are
shown in different colors. The first row are the
bases, and the second row are the attentions. Here $\otimes$  represents element-wise product and $\oplus$  is element-wise sum. Each basis %
multiplies its attention and then is summed to output the final mask.}
\label{fig:blender}
\end{figure}

Recent advances in one-stage object detection prove that one-stage methods such as FCOS can
outperform
their two-stage counterparts in accuracy \cite{tian2019fcos}.
Enabling such one-stage detection frameworks to perform dense instance segmentation is highly desirable as 1)
models consisting of only conventional operations are simpler and easier for cross-platform
deployment; 2) a unified framework provides convenience and flexibility for multi-task network architecture optimization.

Dense instance segmenters can date back to DeepMask~\cite{pinheiro2015learning}, a top-down approach which generates dense instance masks with a sliding window. The representation of mask is encoded into a one-dimensional vector at each spatial location.
Albeit being
simple in structure, it has several obstacles in training that prevent it from achieving superior performance: 1) local-coherence between features and masks is lost;
2) the feature representation is redundant because a mask is repeatedly encoded at each foreground feature; 3) position information is degraded after downsampling with strided convolutions.

The first issue was studied by Dai \etal~\cite{dai2016instance}, who attempt to retain local-coherence by keeping multiple position-sensitive maps. This idea has been explored to its limits by Chen \etal~\cite{chen2019tensormask}, who proposes a dense aligned representation for each location of the target instance mask. However, this approach trades representation efficiency for alignment, making the second issue difficult to resolve. The third issue prevents heavily downsampled features to provide detailed instance information.

Recognizing these difficulties, a line of research takes a bottom-up strategy~\cite{arnab2016bottom,liu2018affinity,davy2019instance}. These methods generate dense per-pixel embedding features and use some techniques to group them. Grouping strategies vary from simple clustering~\cite{de2017semantic} to graph-based algorithms~\cite{liu2018affinity} depending on the embedding characteristics. By performing per-pixel predictions, the local-coherence and position information is well retained. The shortcomings for bottom-up approaches are: 1) heavy reliance on the dense prediction quality, leading to sub-par performance and fragmented/joint masks; 2) limited generalization ability to complex scenes with a large number of classes; 3) requirement for complex post-processing techniques.

In this work, we consider hybridizing top-down and bottom-up approaches. We recognize two important predecessors, FCIS~\cite{li2017fully} and YOLACT~\cite{bolya2019yolact}. They predict instance-level information such as bounding box locations and combine it with per-pixel predictions using
cropping (FCIS) and weighted summation (YOLACT), respectively. We argue that
\textit{these overly simplified assembling designs
may
not provide a good balance for the representation power of top- and bottom-level features.
}

Higher-level features
correspond to
larger receptive field and can better capture overall information about instances such as
poses, while lower-level features preserve better location information and can provide
finer
details. One
of the focuses  of our work is to investigate ways to better merging these two in fully convolutional instance segmentation. More %
specifically,
we generalize the operations for proposal-based mask combination by enriching the instance-level information and performing more fine-grained position-sensitive mask prediction. We carry out extensive ablation studies  to discover the optimal dimensions, resolutions, alignment methods, and feature locations. Concretely, we are able to achieve the followings:

\begin{itemize}

\itemsep -0.1cm

\item We devise a flexible method for proposal-based instance mask generation called blender, which incorporate rich instance-level information with accurate dense pixel features. In head-to-head comparison, our blender surpasses the merging techniques in YOLACT~\cite{bolya2019yolact} and FCIS~\cite{li2017fully} by 1.9 and 1.3 points in mAP on the COCO dataset respectively.

\item We propose a simple architecture, \OurMethod, which is closely tied to the state of the art one-stage object detector, FCOS~\cite{tian2019fcos}, by adding
moldiest computation overhead to the already simple framework.

\item One obvious advantage of \OurMethod is that its inference time does not increase with the number of predictions as conventional two-stage methods do, which makes it more robust in real-time scenarios.

\item The performance of \OurMethod %
achieves mAP of
$37.0\%$  with the ResNet-50~\cite{he2016identity} backbone and $38.4 \%$ mAP with ResNet-101 on the COCO
dataset, outperforming Mask R-CNN~\cite{he2017mask} in accuracy while being about $20\%$ faster. We
set new
records
for fully convolutional instance segmentation, surpassing TensorMask \cite{chen2019tensormask} by 1.1 points in mask mAP with only half training iterations and
$ \nicefrac{1}{5} $
inference time.

To our knowledge, \OurMethod{} may be the first algorithm that
can outperform Mask R-CNN in both mask AP and inference efficiency.

\item
\OurMethod{} can naturally solve panoptic segmentation  without any modification
(refer to Section \ref{sec:Panoptic}),
as the bottom module of \OurMethod{} can segment `\textit{things and stuff}' simultaneously.

\item
    Compared with Mask R-CNN's mask head, which is typically of $ 28 \times 28 $ resolution,   \OurMethod{}'s the bottom module is able to output masks of much higher resolution,
    due to its flexibility and
    the bottom module not being   strictly tied to the FPN.
    Thus \OurMethod{} is able to produce  masks with more accurate edges, as shown in
    Figure~\ref{fig:qualitative}. For applications such as graphics, this can be very important.

\item

    The proposed \OurMethod{} is general and flexible. With minimal modification,
    we can apply \OurMethod{} to solve other instance-level recognition tasks such as keypoint detection.

\end{itemize}

\begin{figure*}[ht]
\centering
\includegraphics[width=.9333859\textwidth]{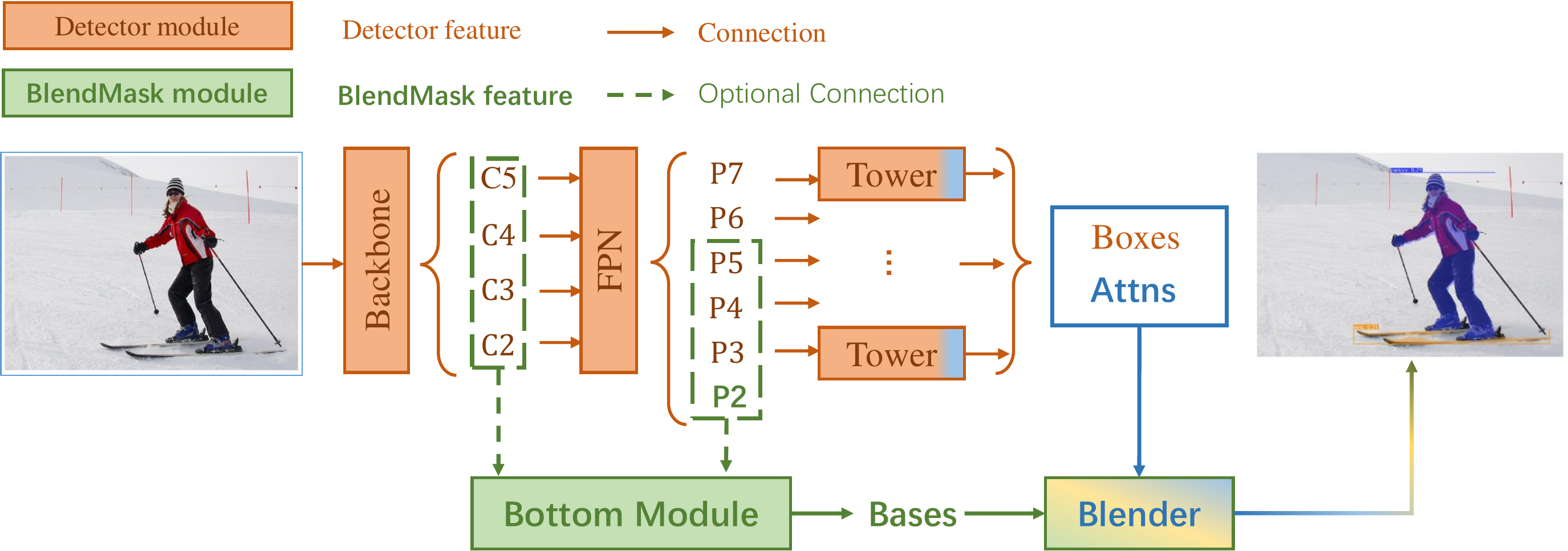}
\caption{\textbf{\OurMethod pipeline}
Our framework
builds upon the state-of-the-art FCOS object detector \cite{tian2019fcos} with
minimal modification.
The bottom module uses either backbone or FPN features to
predict a set of bases. A single convolution layer is added on top of the detection towers to produce attention masks along with each bounding box prediction. For each predicted instance, the blender crops the bases with its bounding box and linearly combine them according the learned attention maps.
Note that the Bottom Module can take features either from `C', or `P' as the input.
}
\label{fig:main}
\end{figure*}

\section{Related work}
\textbf{Anchor-free object detection} Recent advances in object detection unveil the possibilities of removing bounding box anchors
\cite{tian2019fcos}, largely simplifying the detection pipeline. This
much simpler design
improves the box average precision (AP$^{bb}$) by 2.7\%  comparing to its anchor-based counter-part RetinaNet~\cite{lin2017focal}. One possible reason
responsible for
the improvement is that without the restrictions of predefined anchor shapes, targets are freely matched to prediction features according to their effective receptive field.
The hints for us are twofold. First, it is important to map target sizes with proper pyramid levels to fit the effective receptive field for the features. Second, %
{\it removing anchors enables us to assign heavier duties to the top-level instance prediction module  without introducing overall computation overhead}. For example, inferring shape and pose information alongside the bounding box detection would
take about eight times more computation
for anchor-based frameworks than ours.
This
makes it intractable for anchor based detectors to balance the top \textit{vs.}\  bottom workload (i.e., learning instance-aware maps\footnote{Attention maps for \OurMethod and simple weight scalars for YOLACT.}  \vs\  bases).
We assume that this might be the reason why YOLACT
can only learn one single scalar coefficient for each prototype/basis given an instance when computation complexity is taken into account. {\it Only with the use of anchor-free bounding box detectors,  this restriction is removed.
}

\textbf{%
Detect-then-segment
instance segmentation} The dominant instance segmentation paradigms take the two-stage methodology, first detecting the objects and then predicting the foreground masks on each of the proposals. The success of this framework
partially is due to
the alignment operation, RoIAlign~\cite{he2017mask}, which provides local-coherence for the second-stage RoI heads missing in all one-stage top-down approaches. However, two issues exist in two-stage
frameworks. For complicated scenarios with many instances,  inference time for two-stage methods is proportional to the number of instances. Furthermore, the resolution for the RoI features and resulting mask is limited. We %
discuss the second issue in detail in Section~\ref{sec:mask rcnn}.

These problems can be partly solved by replacing a RoI head with a simple crop-and-assemble module. In FCIS, Li \etal~\cite{li2017fully} add a bottom module
to a detection network,
for predicting position-sensitive score maps shared by all instances. This technique was first used in R-FCN~\cite{dai2016r} and later improved in MaskLab~\cite{chen2018masklab}. Each channel of the $k^2$ score maps corresponds to one crop of $k\times k$ evenly partitioned grid tiles of the proposal. Each score map represents the likelihood of the pixel belongs to a object and is at a certain relative position. Naturally, a higher resolution for location crops leads to more accurate predictions, but the computation cost also increases quadratically. Moreover, there are special cases where FCIS representation is not sufficient. When two instances share center positions (or any other relative positions), the score map representation on that crop is ambiguous, it is impossible to tell which instance this crop is describing.

In YOLACT~\cite{bolya2019yolact}, %
an improved
approach is used. Instead of using position-controlled tiles, a set of mask coefficients are learned alongside the box predictions. Then this set of coefficients guides the linear combination of cropped bottom mask bases to generate the final mask. Comparing to FCIS, the responsibility for predicting instance-level information is assigned to the top-level. We argue that using scalar coefficients to encode the instance information is sub-optimal.

To break through these limitations, we propose a
new proposal-based mask generation framework, termed \OurMethod. The top- and bottom-level representation workloads are balanced by a blender module.
Both levels are guaranteed to describe the instance information within their
best capacities.
As shown in our experiments in Section~\ref{sec:experiments}, our blender module improves the performance of bases combination methods comparing to YOLACT and FCIS by a large margin without increasing computation complexity.

\textbf{Refining coarse masks with lower-level features} \OurMethod merges top-level coarse instance information with lower-level fine-granularity.
This idea resembles MaskLab~\cite{chen2018masklab} and Instance Mask Projection (IMP)~\cite{fu2019instance}, which concatenates mask predictions with lower layers of backbone features. The differences are %
clear.
Our coarse mask acts like an attention map. The generation is extremely light-weight, without the need of using semantic or positional supervision, and is closely tied to the object generation. As shown in Section~\ref{sec:bases meanings}, our lower-level features have clear contextual meanings, even though not explicitly guided by bins or crops. Further,  our blender does not require a subnet on top of the merged features as in MaskLab~\cite{chen2018masklab} and IMP~\cite{fu2019instance}, which makes our method more efficient.
In parallel to this work recent  two single shot instance segmentation methods have shown good performance
\cite{PolarMask,wang2019solo}.

\section{Our \OurMethod}
\subsection{Overall pipeline}
\OurMethod consists of a detector network and a mask branch. The mask branch has three parts, a bottom module to predict the score maps, a top layer to predict the instance attentions, and a blender module to merge the scores with attentions. The whole network is illustrated in Figure~\ref{fig:main}.

\textbf{Bottom module} Similar to other proposal-based fully convolutional methods~\cite{li2017fully,bolya2019yolact}, we add a bottom module predicting score maps which we call bases, $\mathbf B$. $\mathbf B$ has a shape of $N\times K\times \frac H s \times \frac W s$, where $N$ is the batch size, $K$ is the number of bases, $H\times W$ is the input size and $s$ is the score map output stride. We use the decoder of DeepLabV3+ in our experiments. Other dense prediction modules should also work without much difference. The input for the bottom module could be backbone features like conventional semantic segmentation networks~\cite{chen2018encoder}, or the feature pyramids like YOLACT and Panoptic FPN~\cite{kirillov2019panoptic}.

\textbf{Top layer} We also append a single convolution layer on each of the detection towers to predict top-level attentions $\mathbf A$. Unlike the mask coefficients in YOLACT, which for each pyramid with resolution $W_l\times H_l$ takes the shape of $N\times K\times H_l\times W_l$, our $\mathbf A$ is a tensor at each location with shape $N\times (K\cdot M\cdot M)\times H_l\times W_l$, where $M\times M$ is the attention resolution. With its 3D structure, our attention map can encode instance-level information, e.g. the coarse shape and pose of the object. $M$ is typically smaller\footnote{The largest $M$ we try is $14$.} than the mask predictions in top-down methods since we only ask for a rough estimate. We predict it with a convolution with $K\cdot M\cdot M$ output channels. Before sending them into the next module, we first apply FCOS~\cite{tian2019fcos} post-process to select the top $D$ box predictions $P = \{\mathbf p_d\in\mathbb R_{\ge 0}^{4}\vert d=1\dots D\}$ and corresponding attentions $A=\{\mathbf a_d\in \mathbb R^{K\times M\times M}\vert d=1\dots D\}$.

\textbf{Blender module} is the key part of our \OurMethod. It combines position-sensitive bases according to the attentions to generate the final prediction. We discuss this module in detail in the next section.

\subsection{Blender module}

The inputs of the blender module are bottom-level bases $\mathbf B$, the selected top-level attentions $A$ and bounding box proposals $P$. First we use RoIPooler in Mask R-CNN~\cite{he2017mask} to crop bases with each proposal $\mathbf p_d$ and then resize the region to a fixed size $R\times R$ feature map $\mathbf r_d$.
\begin{align}
    \mathbf r_d = \mbox{RoIPool}_{R\times R}(\mathbf B, \mathbf p_d),\quad \forall d \in \{1\dots D\}.
\end{align}
More specifically, we use sampling ratio $1$ for RoIAlign, i.e. one bin for each sampling point. The performance of using nearest and bilinear poolers are compared in Table~\ref{table:bottom-align}. During training, we simply use ground truth boxes as the proposals. During inference, we use FCOS prediction results.

Our attention size $M$ is smaller than $R$. We interpolate $\mathbf a_d$ from $M\times M$ to $R\times R$, into the shapes of $R = \{\mathbf r_d\vert d=1\dots D\}$.
\begin{align}
    \mathbf a'_d = \mbox{interpolate}_{M\times M\rightarrow R\times R}(\mathbf a_d),\quad \forall d \in \{1\dots D\}.
\end{align}
Then $\mathbf a'_d$ is normalize with softmax function along the $K$ dimension to make it a set of score maps $\mathbf s_d$.
\begin{align}
    \mathbf s_d = \mbox{softmax}(\mathbf a'_d), \quad \forall d \in \{1\dots D\}.
\end{align}
Then we apply element-wise product between each entity $\mathbf r_d$, $\mathbf s_d$ of the regions $R$ and scores $S$, and sum along the $K$ dimension to get our mask logit $\mathbf m_d$:
\begin{align}
    \mathbf m_d = \sum_{k=1}^K \mathbf s^k_d \circ \mathbf r^k_d, \quad \forall d \in \{1\dots D\},
\end{align}
where $k$ is the index of the basis. We visualize the mask blending process with $K=4$ in Figure~\ref{fig:blender}.

\subsection{Configurations and baselines}\label{sec:configurations}
We consider the following configurable hyper-parameters for \OurMethod:
\begin{itemize}
\itemsep -0.1cm

    \item $R$, the bottom-level RoI resolution,
    \item $M$, the top-level prediction resolution,
    \item $K$, the number of bases,
    \item bottom module input features, it can either be features from the backbone or the FPN,
    \item sampling method for bottom bases, nearest-neighbour or bilinear pooling,
    \item interpolation method for top-level attentions, nearest neighbour or bilinear upsampling.
\end{itemize}
We represent our models with abbreviation \code{R\_K\_M}. For example, \code{28\_4\_4} represents bottom-level region resolution
of
$28\times28$, $4$ number of bases and $4\times4$ top-level instance attentions.
By default, we use backbone features C3 and C5 to keep aligned with DeepLabv3+~\cite{chen2018encoder}. Nearest neighbour interpolation is used in top-level interpolation, for a fair comparison with FCIS~\cite{li2017fully}.
Bilinear sampling is used in the bottom level, consistent with RoIAlign~\cite{he2017mask}.

\begin{figure*}[ht]
  \begin{subfigure}[b]{7cm}
      \begin{subfigure}[b]{0.45\textwidth}
        \includegraphics[width=1\textwidth]{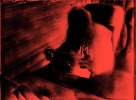}
      \end{subfigure}
      \begin{subfigure}[b]{0.45\textwidth}
        \includegraphics[width=1\textwidth]{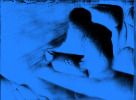}
      \end{subfigure}\\
      \begin{subfigure}[b]{0.45\textwidth}
        \includegraphics[width=1\textwidth]{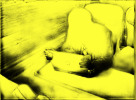}
      \end{subfigure}
      \begin{subfigure}[b]{0.45\textwidth}
        \includegraphics[width=1\textwidth]{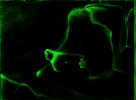}
      \end{subfigure}
      \caption{Bottom-Level Bases}
  \end{subfigure}
  \begin{subfigure}[b]{7cm}
    \includegraphics[width=0.91\textwidth,height=0.6687\textwidth]{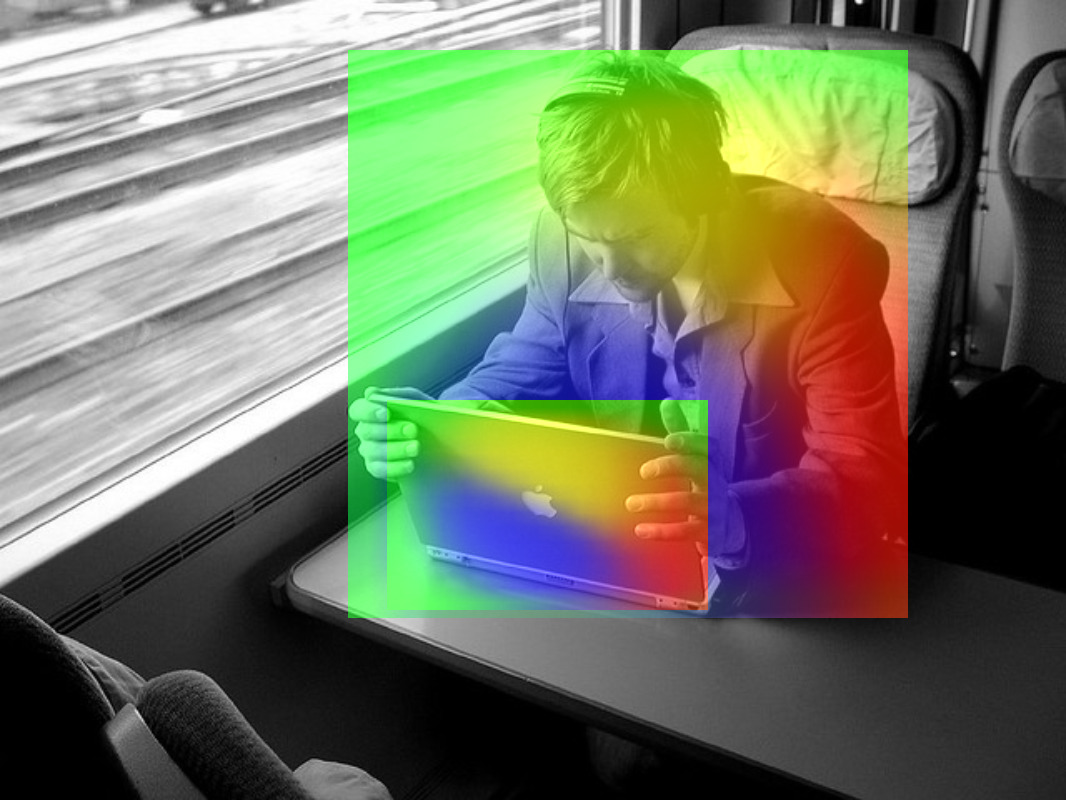}
    \caption{Top-Level attentions}
  \end{subfigure}
\caption{\textbf{Detailed view of learned bases and attentions.} The left four
images are the bottom-level bases. The right
image is the top-level attentions. %
Colors on each position of the attentions correspond to the weights of the bases,
indicating from which part of which base is the mask assembled.}
\label{fig:blend}
\end{figure*}

\subsection{
Semantics %
encoded in
learned bases and attentions}\label{sec:bases meanings}

By examining the generated bases and attentions on \code{val2017}, we observe this pattern. On its bases, \OurMethod encodes two types of local information, 1) whether the pixel is on an object (semantic masks), 2) whether the pixel is on certain part of the object (position-sensitive features).

The complete bases and attentions projected onto the original image are illustrated in Figure~\ref{fig:blend}. The first two bases (red and blue) detects points on the upper-right and bottom-left parts of the objects. The third (yellow) base activates on points more likely to be on an object. The fourth (green) base only activates on the borders of objects. Position-sensitive features help us separate overlapping instances, which enables \OurMethod to represent all instances more efficiently than YOLACT~\cite{bolya2019yolact}. The positive semantic mask makes our final prediction smoother than FCIS~\cite{li2017fully} and the negative one can further suppress out-of-instance activations. We compare our blender with YOLACT and FCIS counterparts in Table~\ref{table:merging}. \OurMethod can learn more accurate features than YOLACT and FCIS with much fewer number of bases (4 \vs\ 32 \vs\  49, see Section~\ref{sec:qualitative}).

\section{Experiments}\label{sec:experiments}
Our experiments are reported on the MSCOCO 2017 instance segmentation datatset~\cite{lin2014microsoft}. It contains 123K images with 80-class instance labels. Our models are trained on the \code{train2017} split (115K images) and the ablation study is carried out on the \code{val2017} split (5K images). Final results are on \code{test}-\code{dev}.
The evaluation metrics are COCO mask average precision (AP), AP at IoU 0.5 (AP$_{50}$), 0.75 (AP$_{75}$) and AP for objects at different sizes AP$_S$, AP$_M$, and AP$_L$.

\textbf{Training details} Unless specified, ImageNet pre-trained ResNet-50~\cite{He2015Deep} is used as our backbone network. DeepLabv3+~\cite{chen2018encoder} with channel width $128$ is used as our bottom module. For ablation study, all the networks are trained with the $1\times$ schedule of FCOS~\cite{tian2019fcos}, i.e., 90K iterations, batch size 16 on 4 GPUs, and base learning rate 0.01 with constant warm-up of 1k iterations. The learning rate is reduced by a factor of 10 at iteration 60K and 80K. Input images are resized to have shorter side 800 and longer side at maximum 1333. All hyperparameters are set to be the same with FCOS~\cite{tian2019fcos}.

\textbf{Testing details} The unit for inference time is `ms' in all our tables. For the ablation experiments, performance and time of our models are measured with one image per batch on one 1080Ti GPU.

\subsection{Ablation experiments}\label{sec:ablation}
We investigate the effectiveness of our blender module by carrying out ablation experiments on the configurable hyperparameters in Section~\ref{sec:configurations}.

\begin{table}[t]
\small
\centering
\begin{tabular}{r|ccc}
\hline
Method & AP & AP$_{50}$ & AP$_{75}$ \\
\hline
\hline
Weighted-sum & 29.7   &  52.2      &  30.1    \\
Assembler & 30.3   &  52.5    & 31.3    \\
Blender & \textbf{31.6}   &   \textbf{53.4}     &  \textbf{33.3 }   \\
\hline
\end{tabular}
\caption{
{\bf
Comparison of different  strategies
for merging
top and bottom modules. }
Here the model used is \code{28\_4\_4}.
Wei\-ght\-ed-\-\-sum is our analogy to YOLACT, reducing the top resolution to $1 \times 1$. Assembler is our analogy to FCIS,
where the number of bases is increased to $16$, matching each of the region crops without the need of top-level attentions.}
\label{table:merging}
\end{table}

\textbf{Merging methods: Blender  \vs\  YOLACT \vs\  FCIS} Similar to our method, YOLACT~\cite{bolya2019yolact} and FCIS~\cite{li2017fully} both merge proposal-based bottom regions to create mask prediction. YOLACT simply performs a weighted sum of the channels of the bottom regions; FCIS assembles crops of position-sensitive masks without modifications. Our blender can be regarded as a generalization where both YOLACT and FCIS merging are special cases: The blender with $1\times1$ top-level resolution degenerates to YOLACT; and FCIS is the case where we use fixed one-hot blending attentions and nearest neighbour top-level interpolation.

Results of these variations are shown in Table~\ref{table:merging}. Our blender surpasses the other alternatives by a large margin. We assume the reason is that other methods lack instance-aware guidance on the top. By contrast, our blender has a fine-grained top-level attention map, as illustrated in Figure~\ref{fig:blend}.

\begin{table}[t!]
\centering
\small
\begin{tabular}{c|c|c|c|ccc}
\hline
$R$ & $M$ & Time & AP & AP$_S$ & AP$_M$ & AP$_L$ \\
\hline
\hline
\multirow{3}{*}{$28$} & $2$ & \textbf{72.7} & 30.6 & 14.3 & 34.1 & 42.5 \\
&$4$                   & 72.9 & 31.6  & 14.8 & 35.2 & 45.0 \\
&$7$                   & 73.9 & 32.0  & 15.3  & 35.6  & 45.0  \\
\hline
\multirow{3}{*}{$56$} &$4$  & 72.9  & 32.5  & 14.9 & 36.1 & 46.0    \\
&$7$                  & 74.1 & 33.1   & 15.1 & 36.6 & \textbf{47.7} \\
&$14$                & 77.7 & \textbf{33.3} & \textbf{16.3} & \textbf{36.8} & 47.4 \\
\hline
\end{tabular}
\caption{\textbf{Resolutions}: Performance %
by
varying
top-/bottom-level resolutions, with
the number of bases $K=4$ for all models.
Top-level attentions are interpolated with nearest neighbour. Bottom module uses backbone features C3, C5. The performance increases as the attention resolution grows, saturating at resolutions of near $1/4$ of the region sizes.
}
\label{table:crops}
\end{table}

\textbf{Top and bottom resolutions}: We measure the performances of our model with different top- and bottom-level resolutions, trying bottom pooler resolution $R$ being $28$ and $56$, with $R/M$ ratio from $14$ to $4$. As shown in Table~\ref{table:crops}, by increasing the attention resolution, we can incorporate more detailed instance-level information while keeping the running time roughly the same. Notice that the gain slows down at higher resolutions revealing limit of detailed information on the top-level. So we don't include larger top settings with $R/M$ ratio smaller than $4$.

Different from two-stage approaches, increasing the bottom-level bases pooling resolution does not introduce much computation overhead. Increasing it from $28$ to $56$ only increases the inference time within 0.2ms while mask AP increases by 1 point. In further ablation experiment, we set $R=56$ and $M=7$ for our baseline model if not specified.

\begin{table}[b]
\small
\centering
\begin{tabular}{c|ccc}
\hline
$K$ & AP & AP$_{50}$ & AP$_{75}$ \\
\hline
\hline
1 & 30.6   &  52.9    & 31.6   \\
2 & 31.2   &  53.4    & 32.3    \\
4 & \textbf{33.1}   &   \textbf{54.1}     &  \textbf{34.9}    \\
8 & 33.0   &   53.9     &  \textbf{34.9}    \\
\hline
\end{tabular}
\caption{\textbf{Number of bases}: Performances of \code{56\_K\_7} models.
For the configuration of one basis,
we use sigmoid activation for both top and bottom features.
Our model works with a small number of
bases.}
\label{table:bases}
\end{table}

\textbf{Number of bases}: YOLACT~\cite{bolya2019yolact} uses $32$ bases concerning the inference time. With our blender, the number of bases can be further reduced, to even just one. We report our models with number of bases varying from $1$ to $8$. Different from normal blender, the one-basis version uses sigmoid activation on both the base and the attention map. Results are shown in Table~\ref{table:bases}. Since instance-level information is better represented with the top-level attentions, we only need $4$ bases to get the optimal accuracy. $K=4$ is adopted by all
subsequent
experiments.

\begin{table}[ht]
\small
\centering
\begin{tabular}{ r |c|c|ccc}
\hline
Features & $M$ & Time (ms) & AP & AP$_{50}$ & AP$_{75}$ \\
\hline
\hline
\multirow{2}{*}{C3, C5} & $7$ & 74.1 & 33.1   &  54.1    & 34.9  \\
&$14$                   & 77.7 & 33.3 & 54.1  & 35.3 \\
\hline
\multirow{2}{*}{P3, P5} & $7$   & \textbf{72.5} &  33.3  &  54.2 & 35.3\\
&$14$                & 76.4 & \textbf{33.4} &  \textbf{54.3} & \textbf{35.5} \\
\hline
\end{tabular}
\caption{\textbf{Bottom feature locations}: Performance with bottom resolution $56\times56$, $4$ bases and bilinear bottom interpolation. C3, C5 uses features from backbone. P3, P5 uses features from FPN.}
\label{table:fpn}
\end{table}

\textbf{Bottom feature locations: backbone \vs\  FPN} We compare our bottom module feature sampling locations. By using FPN features, we can improve the performance while reducing the running time (see Table~\ref{table:fpn}). In later experiments, if not specified, we use P3 and P5 of FPN as our bottom module input.

\begin{table}[h]
\small
\centering
\begin{tabular}{ r |c|ccc}
\hline
Interpolation & $M$ & AP & AP$_{50}$ & AP$_{75}$ \\
\hline
\hline
\multirow{2}{*}{Nearest} & $7$ &  33.3  & 54.2   &   35.3 \\
&$14$                   & 33.4  & 54.3  & 35.5   \\
\hline
\multirow{2}{*}{Bilinear} &$7$   & 33.5  & 54.3  & \textbf{35.7}  \\
&$14$      & \textbf{33.6} & \textbf{54.6}   & 35.6    \\
\hline
\end{tabular}
\caption{\textbf{Top interpolation}: Performance with bottom resolution $56\times56$, $4$ bases and bilinear bottom interpolation. Nearest represents nearest-neighbour upsampling and bilinear is bilinear interpolation.}
\label{table:top-align}
\end{table}

\begin{table}[h]
\small
\centering
\begin{tabular}{ r |c|c|ccc}
\hline
Alignment                   & $R$  & $M$ &  AP   & AP$_{50}$ & AP$_{75}$ \\
\hline
\hline
\multirow{2}{*}{Nearest}    &$28$  & $7$    & 30.5  & 53.0  & 31.6  \\
                            &$56$   & $14$  & 31.9  & 53.6  & 33.4   \\
\hline
\multirow{2}{*}{Bilinear}   &$28$   & $7$   & 32.4 & 54.4 & 34.5 \\
                            &$56$   & $14$  & \textbf{33.6} & \textbf{54.6} & \textbf{35.6}\\
\hline
\end{tabular}
\caption{\textbf{Bottom Alignment}: Performance with $4$ bases and bilinear top interpolation. Nearest represents the original RoIPool in Fast R-CNN~\cite{girshick2015fast} and bilinear is the RoIAlign in Mask R-CNN~\cite{he2017mask}.}
\label{table:bottom-align}
\end{table}

\begin{table}[ht]
\centering
\small
\begin{tabular}{ r |c|c|ccc}
\hline
Bottom & Time (ms) & AP$^{bb}$ & AP & AP$_{50}$ & AP$_{75}$ \\
\hline
\hline
DeepLabV3+  & \textbf{76.5} & 38.8 & 33.6 & 54.6   & 35.6 \\
+semantic & \textbf{76.5} & \textbf{39.2} & 34.2 & 54.9  & 36.4   \\
+128     & 78.5 & 39.1 & 34.3 & 54.9  & 36.6   \\
+s/4      & 86.4 & \textbf{39.2} & \textbf{34.4}  & \textbf{55.0} & \textbf{36.8}    \\
\hline
Proto-P3 & 85.2 & 39.0 & \textbf{34.4} & 54.9 & \textbf{36.8} \\
Proto-FPN & 78.8 & 39.1 & \textbf{34.4} & 54.9   & \textbf{36.8} \\
\hline
\end{tabular}
\caption{\textbf{Other improvements}: We use \code{56\_4\_14x14} with bilinear interpolation for all models. `+semantic' is the model with semantic supervision as auxiliary loss. `+128' is the model with bottom module channel size being 256. `+s/4' means using P2,P5 as the bottom input. Decoders in DeepLab V3+ %
and YOLACT (Proto) are compared. `Proto-P3' has channel width of 256 and `Proto-FPN' of 128. Both are trained with `+semantic' setting.}
\label{table:improvements}
\end{table}

\begin{table*}[ht]
\centering
\small
\begin{tabular}{r |c|c|c|c|ccc|ccc}
\hline
Method &Backbone &Epochs &Aug.\ & Time (ms) & AP &AP$_{50}$ &AP$_{75}$ &AP$_S$ &AP$_M$ &AP$_L$\\
\hline
\hline
Mask R-CNN\cite{he2017mask}&\multirow{5}{*}{R-50}&12& & 97.0 & 34.6 & 56.5 & 36.6 & 15.4 & 36.3 & 49.7 \\
Mask R-CNN*& &72&\checkmark & 97+& 36.8 & \textbf{59.2} & 39.3 & 17.1 & 38.7 & 52.1 \\
TensorMask\cite{chen2019tensormask}& &72&\checkmark& 400+ & 35.5 &57.3&37.4&16.6&37.0&49.1\\
\OurMethod&&12& & 78.5 & 34.3 & 55.4 & 36.6 & 14.9 & 36.4 & 48.9\\
\OurMethod& &36&\checkmark & 78.5 & 37.0 & 58.9 & 39.7 & 17.3 & 39.4 & 52.5\\
\OurMethod{}*& &36&\checkmark & \textbf{74.0} & \textbf{37.8} & 58.8 & \textbf{40.3} & \textbf{18.8} & \textbf{40.9} & \textbf{53.6}\\
\hline
Mask R-CNN &\multirow{8}{*}{R-101}& 12&  & 118.1 &36.2 & 58.6 & 38.4 & 16.4 & 38.4 & 52.1\\
Mask R-CNN* &  & 36 &\checkmark& 118+& 38.3 & 61.2 & 40.8 & 18.2 &40.6&54.1\\
TensorMask &  & 72&\checkmark & 400+ & 37.3 &59.5 &39.5&17.5 &39.3&51.6\\
SOLO~\cite{wang2019solo} & & 72 & \checkmark & - & 37.8 & 59.5 & 40.4 & 16.4 & 40.6 & 54.2 \\
+deform convs
~\cite{wang2019solo} & & 72 & \checkmark & - & 40.4 & 62.7 & 43.3 & 17.6 & 43.3 & 58.9 \\
\OurMethod&  &36&\checkmark & 101.8 & 38.4 & 60.7& 41.3 & 18.2 & 41.5 & 53.3 \\
\OurMethod{}*&  &36&\checkmark & \textbf{94.1} & 39.6 & 61.6& 42.6 & 22.4 & 42.2 & 51.4 \\
+deform convs (interval $=3$)& &60&\checkmark & 105.0 & \textbf{41.3} & \textbf{63.1}& \textbf{44.6} & \textbf{22.7} & \textbf{44.1} & \textbf{54.5} \\
\hline
\end{tabular}
\caption{\textbf{Quantitative results} on COCO \code{test}-\code{dev}. We compare our \OurMethod %
against Mask R-CNN and TensorMask. Mask R-CNN* is the modified Mask R-CNN with implementation details in TensorMask~\cite{chen2019tensormask}. Models with `aug.' uses multi-scale training with shorter side range $[640, 800]$. Speed for Mask R-CNN 1$\times$ and \OurMethod are measured with \code{maskrcnn\_benchmark} on a single 1080Ti GPU. \OurMethod{}* is implemented with \code{Detectron2}, the speed difference is caused by different measuring rules. `+deform convs (interval $=3$)' uses deformable convolution in the backbone with interval 3, following~\cite{bolya2019yolact++}.}
\label{table:main}
\end{table*}

\begin{table*}[h]
\small
\centering
\begin{tabular}{ r |c|c|c|c|c|ccc}
\hline
Method                   & Backbone & NMS & Resolution  & Time (ms) & AP$^{bb}$ & AP &AP$_{50}$ &AP$_{75}$ \\
\hline
\hline
YOLACT    &\multirow{3}{*}{R-101} &Fast & $550\times550$ & \textbf{34.2}  & 32.5 & 29.8 & 48.3 & 31.3 \\
YOLACT & &Fast & $700\times700$  & 46.7  & 33.4 & 30.9 & 49.8 & 32.5  \\
\OurMethod-RT & & Batched & $550\times *$  & 47.6 & \textbf{41.6} & \textbf{36.8} & \textbf{61.2} & \textbf{42.4}  \\
                            
\hline
Mask R-CNN   &\multirow{2}{*}{R-50}   &\multirow{2}{*}{Batched} & \multirow{2}{*}{$550\times *$}   & 63.4 & 39.1 & \textbf{35.3} & \textbf{56.5} & \textbf{37.6} \\
\OurMethod-RT & &  &  & \textbf{36.0} & \textbf{39.3} & 35.1 & 55.5 & 37.1 \\
\hline
\end{tabular}
\caption{\textbf{Real-time setting comparison} of speed and accuracy with other state-of-the-art methods on COCO \code{val2017}. Metrics for YOLACT are obtained using their official code and trained model. Mask R-CNN and \OurMethod models are trained and measured using \code{Detectron2}. Resolution $550\times *$ means using shorter side $550$ in inference.
Our fast version of \OurMethod{} significantly outperforms YOLACT in accuracy with 
\textit{on par} execution time. 
}
\label{table:real-time}
\end{table*}

\textbf{Interpolation method: nearest \vs\  bilinear} In Mask R-CNN~\cite{he2017mask}, RoIAlign plays a crucial role in aligning the pooled features to keep local-coherence. We investigate the effectiveness of bilinear interpolation for bottom RoI sampling and top-level attention re-scaling. As shown in Table~\ref{table:top-align}, changing top interpolation from nearest to bilinear yields a marginal improvement of 0.2 AP.

The results of bottom sampling with RoIPool~\cite{girshick2015fast} (nearest) and RoIAlign~\cite{he2017mask} (bilinear) are shown in Table~\ref{table:bottom-align}. For both resolutions, the aligned bilinear sampling could improve the performance by almost 2AP. Using aligned features for the bottom-level is more crucial, since it is where the detailed positions are predicted. Bilinear top and bottom interpolation are adopted for our final models.

\textbf{Other improvements}: We experiment on other tricks to improve the performance. First we add auxiliary semantic segmentation supervision on P3 similar to YOLACT~\cite{bolya2019yolact}. Then we increase the width of our bottom module from $128$ to $256$. Finally, we reduce the bases output stride from $8$ to $4$, to produce higher-quality bases. We achieve this by using P2 and P5 as the bottom module input. Table~\ref{table:improvements} shows the results. By adding semantic loss, detection and segmentation results are both improved. This is an interesting effect since the instance segmentation task itself does not improve the box AP. Although all tricks contribute to the improvements,  we decide to %
not
use larger basis resolution because it slows down the model by 10ms per image.

We also implement the protonet module in YOLACT~\cite{bolya2019yolact} for comparison. We include a P3 version and an FPN version. The P3 version is identical to the one used in YOLACT. For the FPN version, we first change the channel width of P3, P4, and P5 to 128 with a $3\times3$ convolution. Then upsample all features to s/8 and sum them up. Following are the same as P2 version except that we reduce convolution layers by one. Auxiliary semantic loss is applied to both versions. As shown in Table~\ref{table:improvements}, changing the bottom module from DeepLabv3+ to protonet does not modify the speed and performance significantly.

\begin{figure*}[ht]
{
  \begin{subfigure}[b]{7.5cm}
      \begin{subfigure}[b]{3cm}
        \includegraphics[width=3cm]{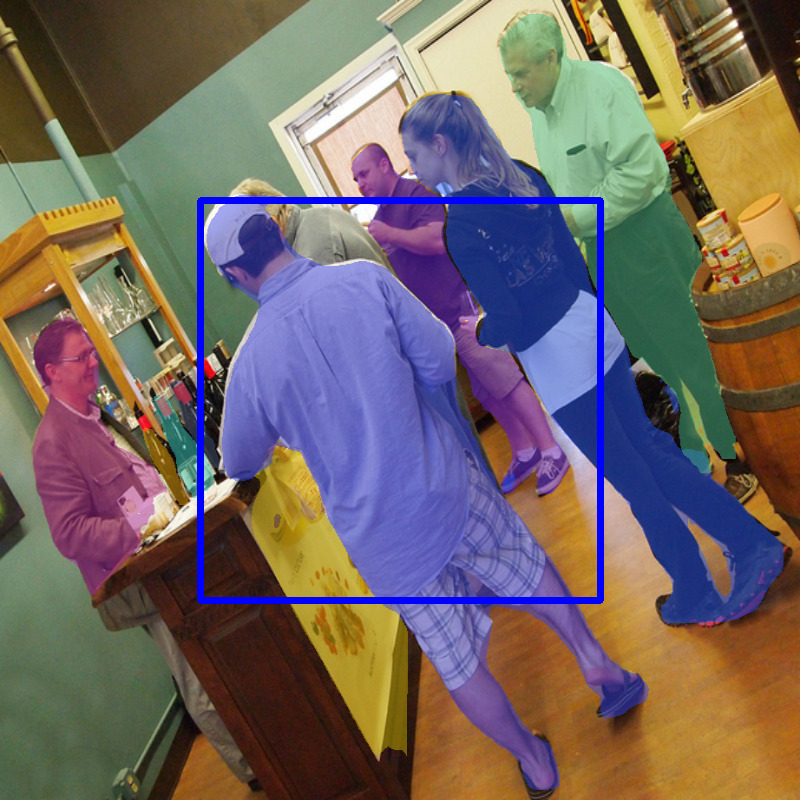}
        \caption*{ours}
      \end{subfigure}%
      \begin{subfigure}[b]{1.5cm}
        \begin{subfigure}[b]{1.5cm}
          \includegraphics[width=1.5cm]{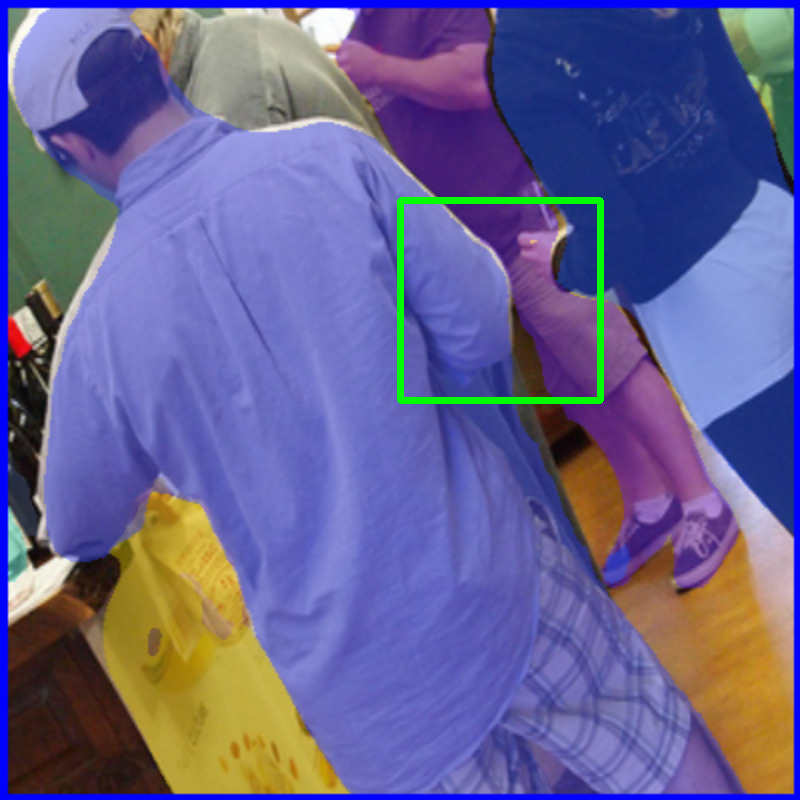}
        \end{subfigure}%
        \\[-0.2ex]
        \begin{subfigure}[b]{1.5cm}
          \includegraphics[width=1.5cm]{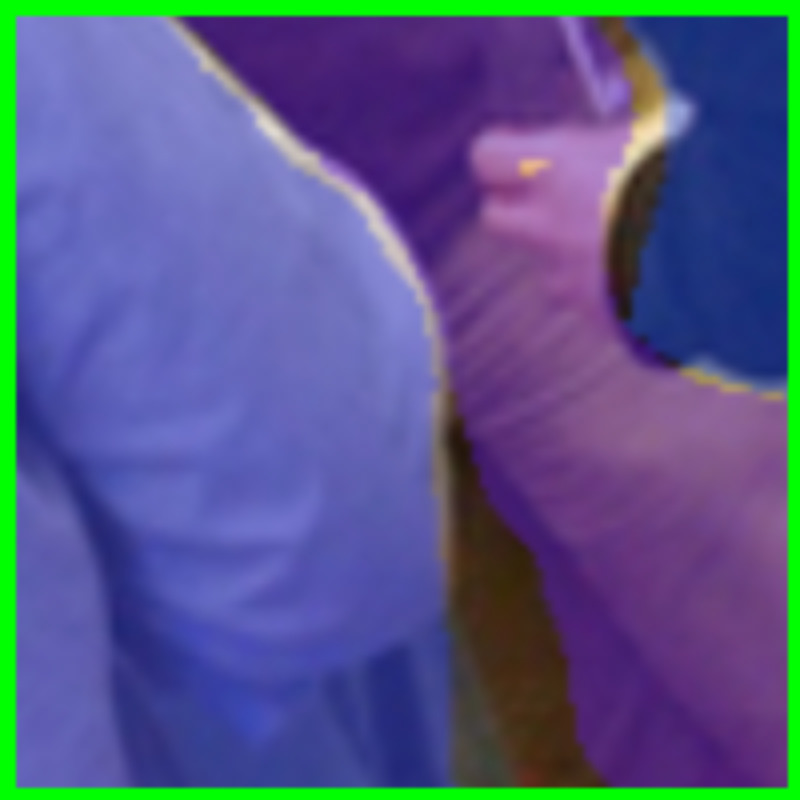}
        \end{subfigure}%
      \caption*{ours}
      \end{subfigure}%
      \begin{subfigure}[b]{1.5cm}
        \begin{subfigure}[b]{1.5cm}
          \includegraphics[width=1.5cm]{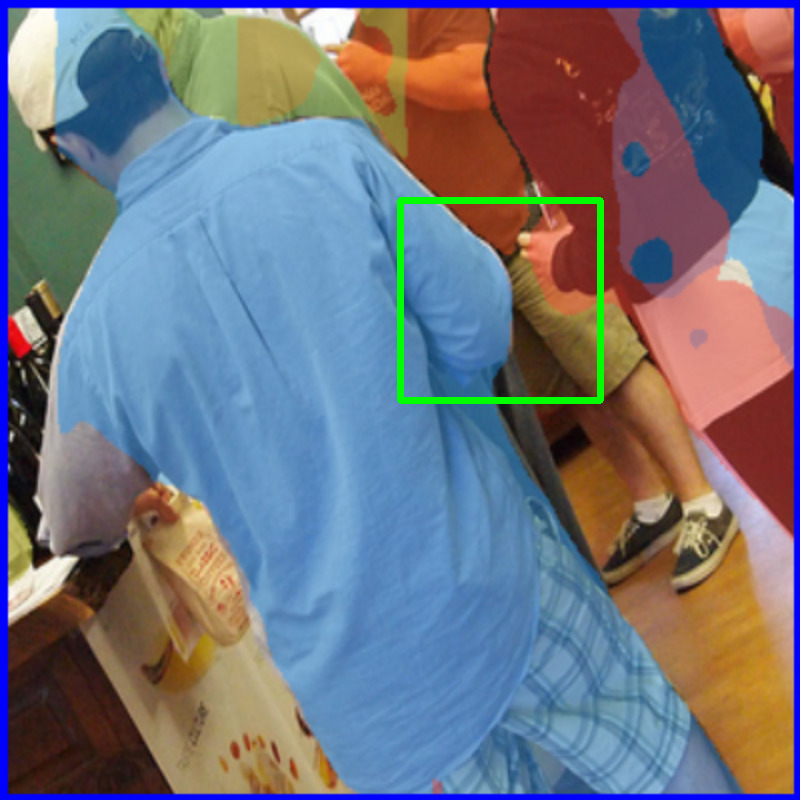}
        \end{subfigure}%
        \\[-0.2ex]
        \begin{subfigure}[b]{1.5cm}
          \includegraphics[width=1.5cm]{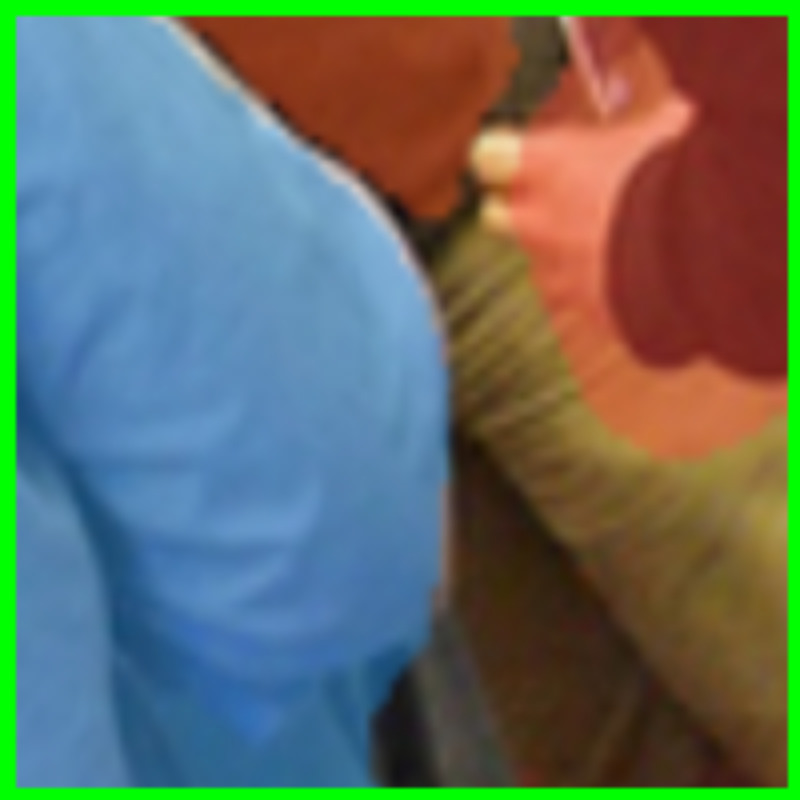}
        \end{subfigure}%
      \caption*{YOLACT}
      \end{subfigure}%
      \begin{subfigure}[b]{1.5cm}
        \begin{subfigure}[b]{1.5cm}
          \includegraphics[width=1.5cm]{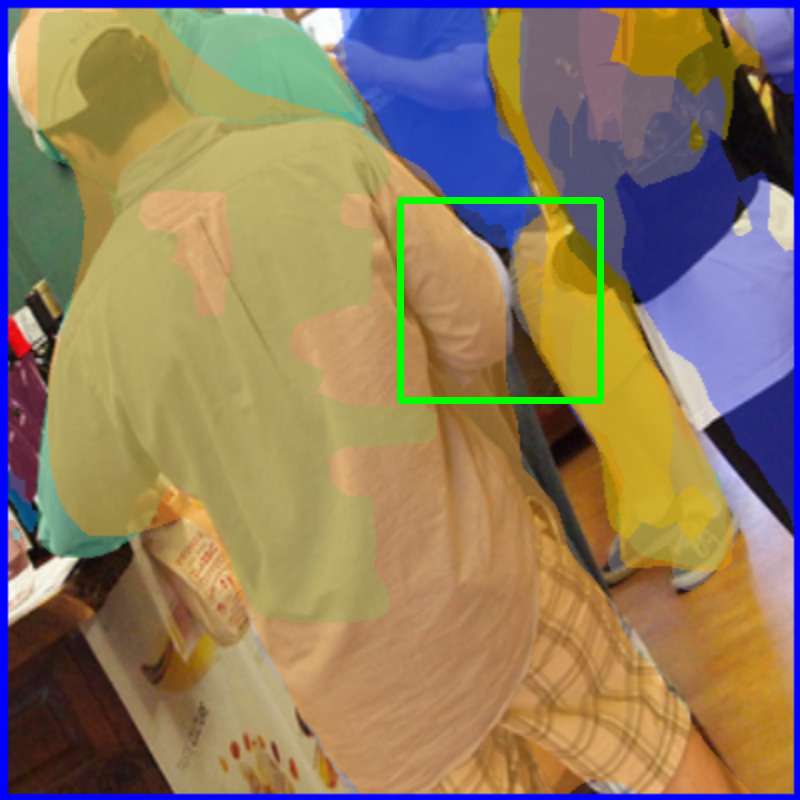}
        \end{subfigure}%
        \\[-0.2ex]
        \begin{subfigure}[b]{1.5cm}
          \includegraphics[width=1.5cm]{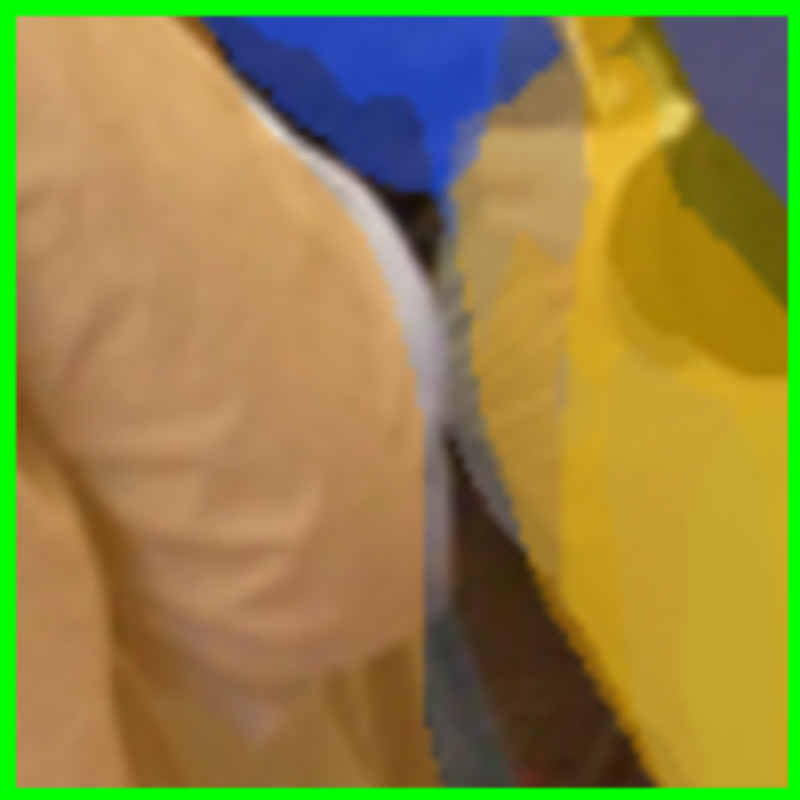}
        \end{subfigure}%
      \caption*{M-RCNN}
      \end{subfigure}%
  \end{subfigure}
  \begin{subfigure}[b]{7.5cm}
      \begin{subfigure}[b]{3cm}
        \includegraphics[width=3cm]{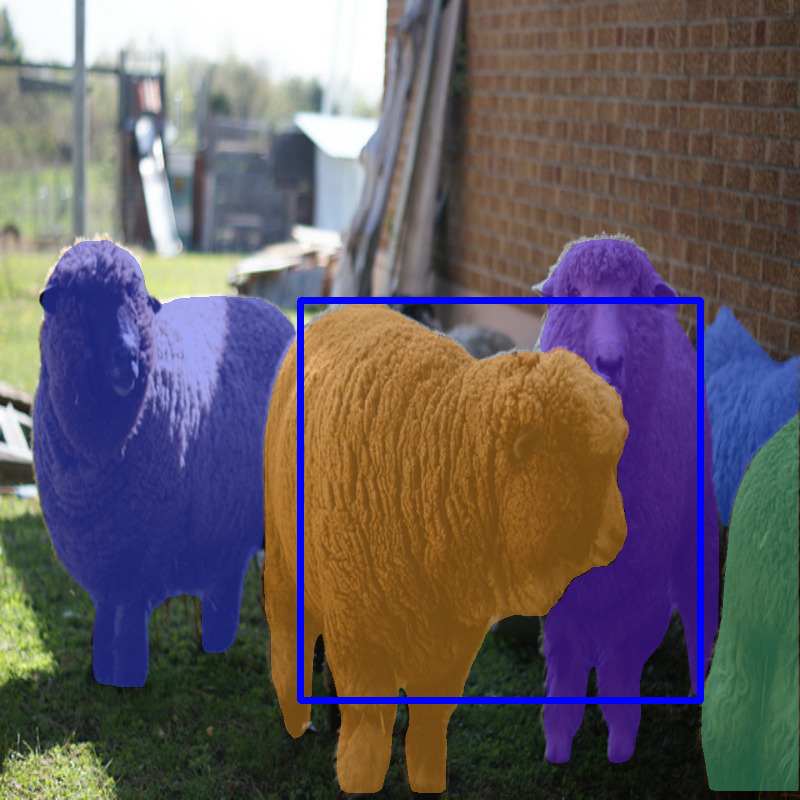}
        \caption*{ours}
      \end{subfigure}%
      \begin{subfigure}[b]{1.5cm}
        \begin{subfigure}[b]{1.5cm}
          \includegraphics[width=1.5cm]{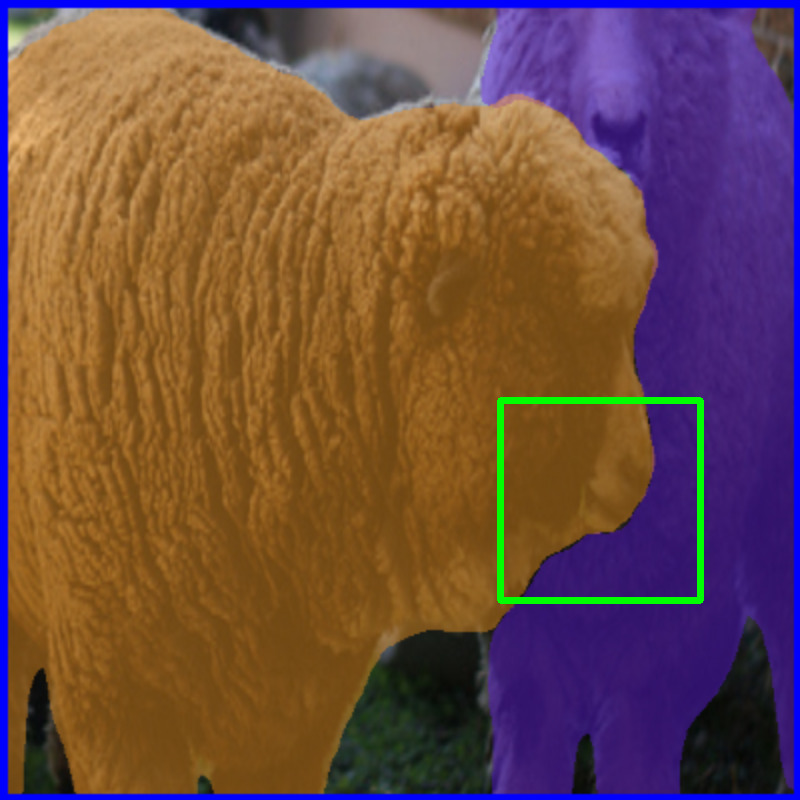}
        \end{subfigure}%
        \\[-0.2ex]
        \begin{subfigure}[b]{1.5cm}
          \includegraphics[width=1.5cm]{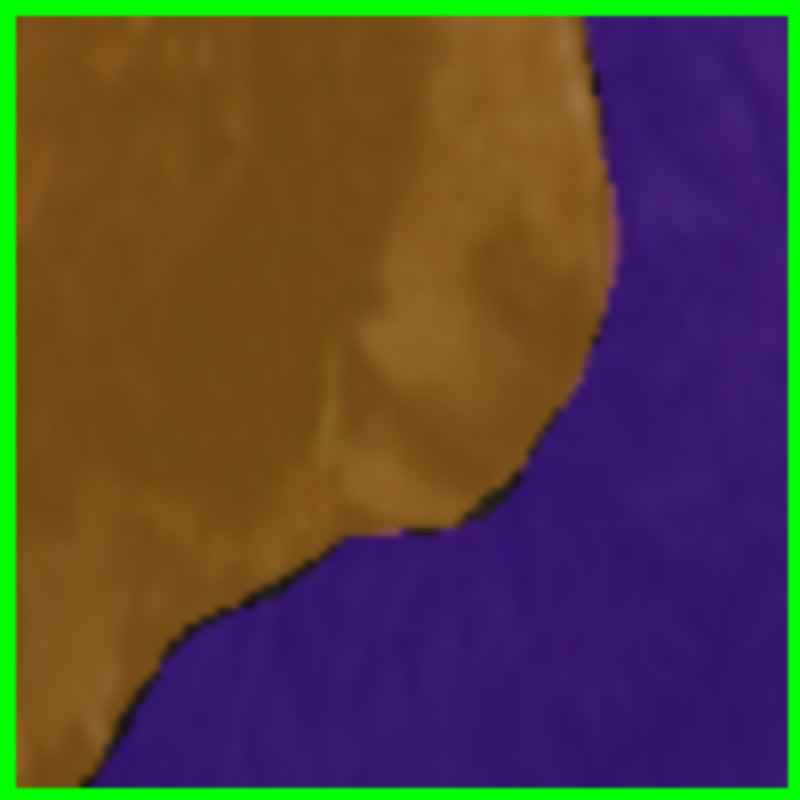}
        \end{subfigure}%
        \caption*{ours}
      \end{subfigure}%
      \begin{subfigure}[b]{1.5cm}
        \begin{subfigure}[b]{1.5cm}
          \includegraphics[width=1.5cm]{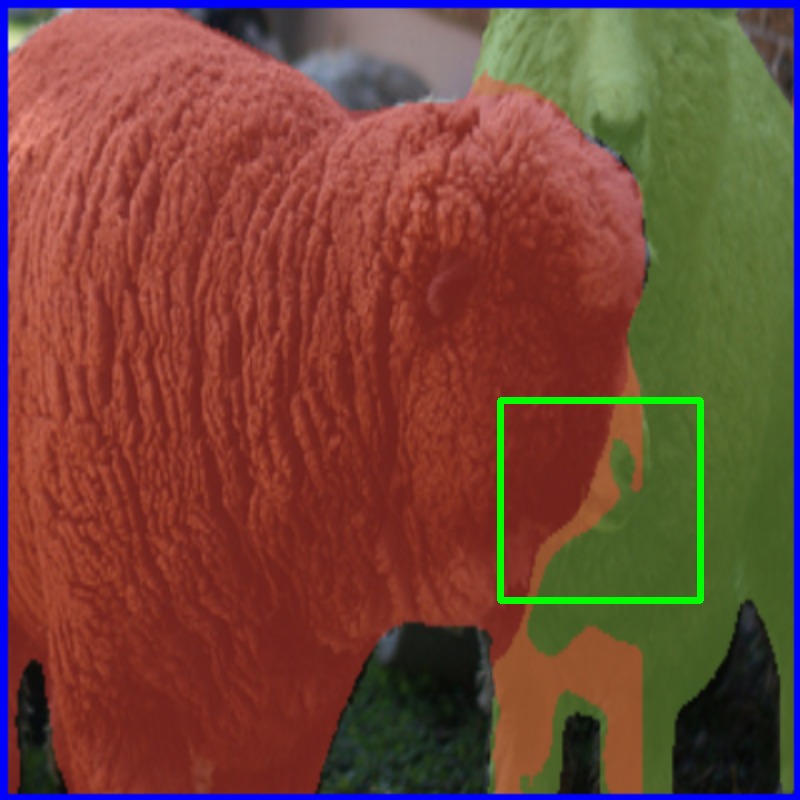}
        \end{subfigure}%
        \\[-0.2ex]
        \begin{subfigure}[b]{1.5cm}
          \includegraphics[width=1.5cm]{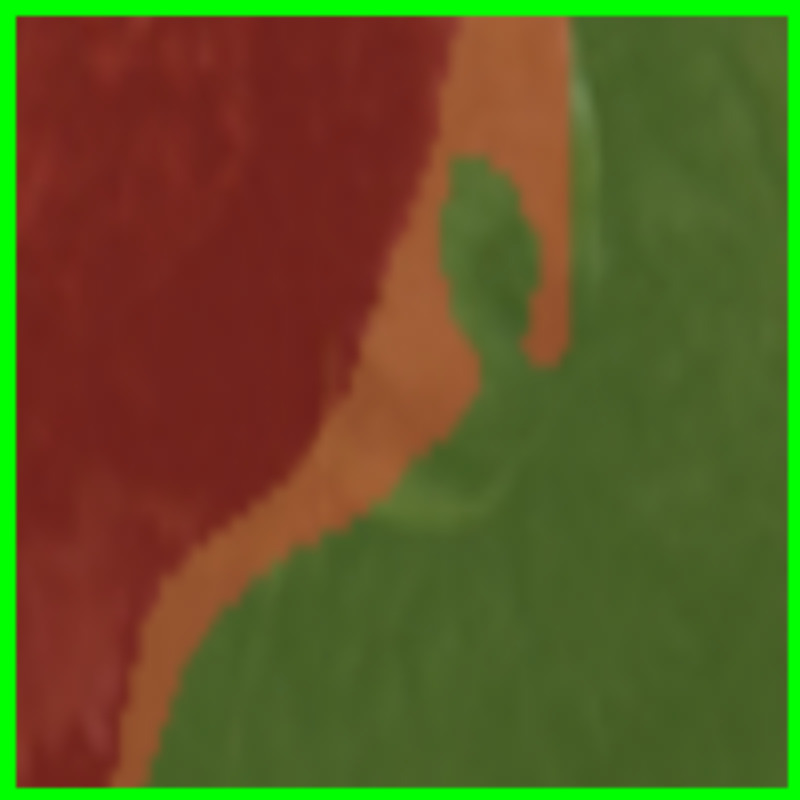}
        \end{subfigure}%
        \caption*{YOLACT}
      \end{subfigure}%
      \begin{subfigure}[b]{1.5cm}
        \begin{subfigure}[b]{1.5cm}
          \includegraphics[width=1.5cm]{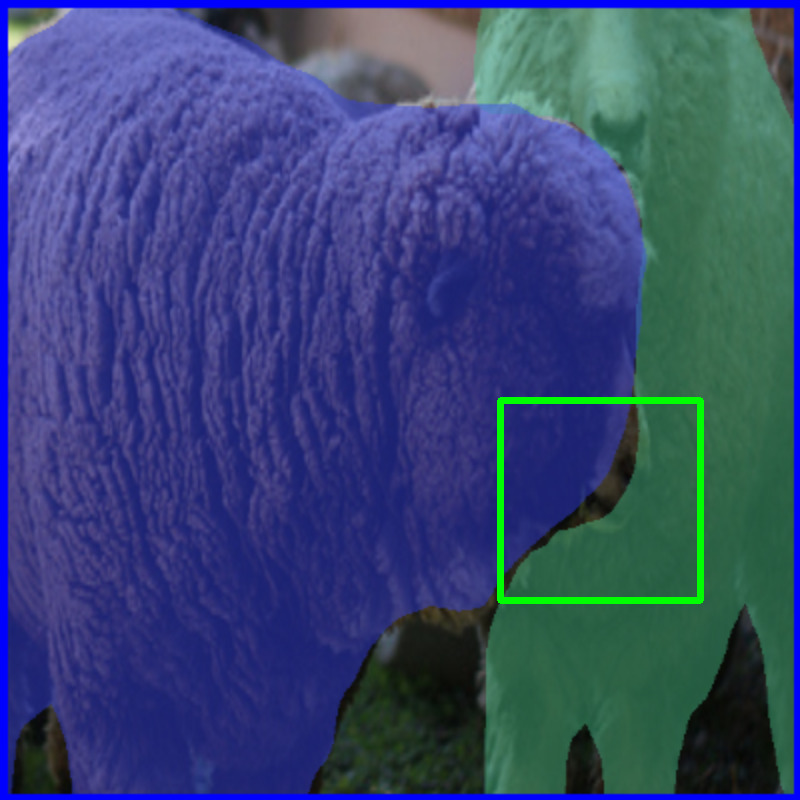}
        \end{subfigure}%
        \\[-0.2ex]
        \begin{subfigure}[b]{1.5cm}
          \includegraphics[width=1.5cm]{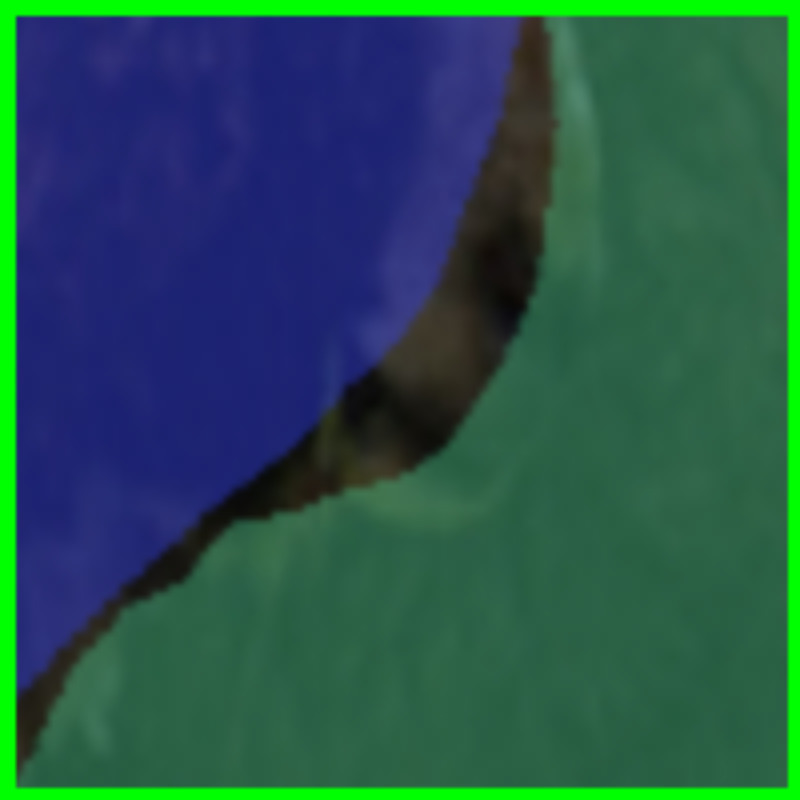}
        \end{subfigure}%
        \caption*{M-RCNN}
      \end{subfigure}%
  \end{subfigure}\\
  \begin{subfigure}[b]{7.5cm}
      \begin{subfigure}[b]{3cm}
        \includegraphics[width=3cm]{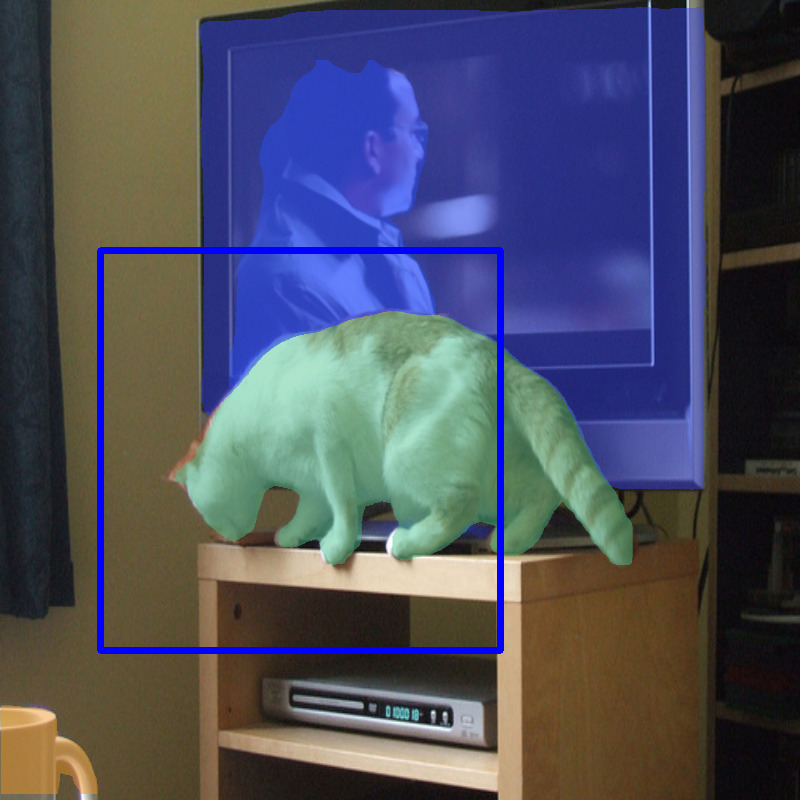}
        \caption*{ours}
      \end{subfigure}%
      \begin{subfigure}[b]{1.5cm}
        \begin{subfigure}[b]{1.5cm}
          \includegraphics[width=1.5cm]{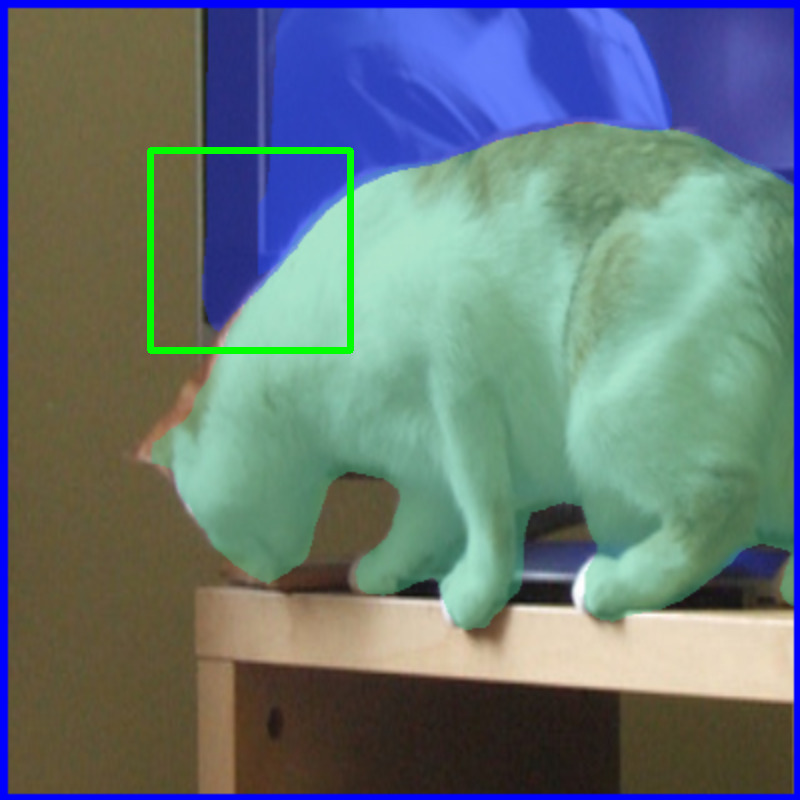}
        \end{subfigure}%
        \\[-0.2ex]
        \begin{subfigure}[b]{1.5cm}
          \includegraphics[width=1.5cm]{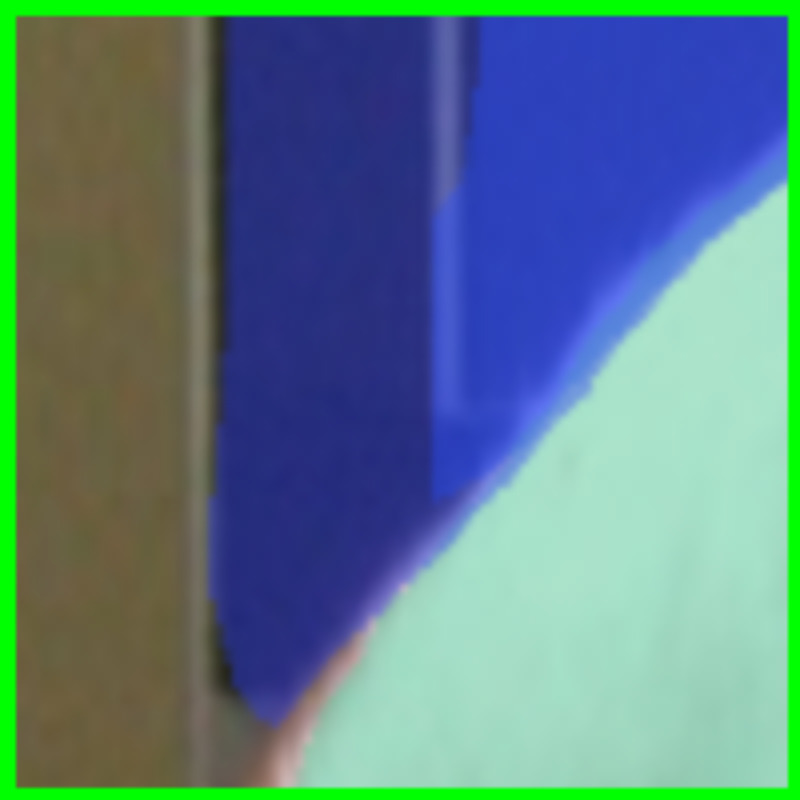}
        \end{subfigure}%
        \caption*{ours}
      \end{subfigure}%
      \begin{subfigure}[b]{1.5cm}
        \begin{subfigure}[b]{1.5cm}
          \includegraphics[width=1.5cm]{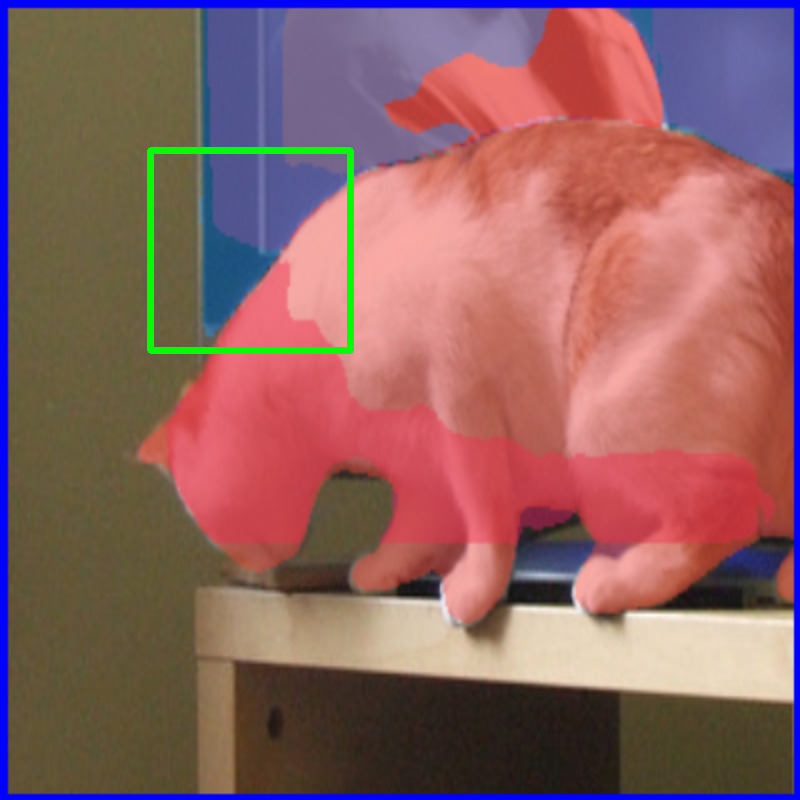}
        \end{subfigure}%
        \\[-0.2ex]
        \begin{subfigure}[b]{1.5cm}
          \includegraphics[width=1.5cm]{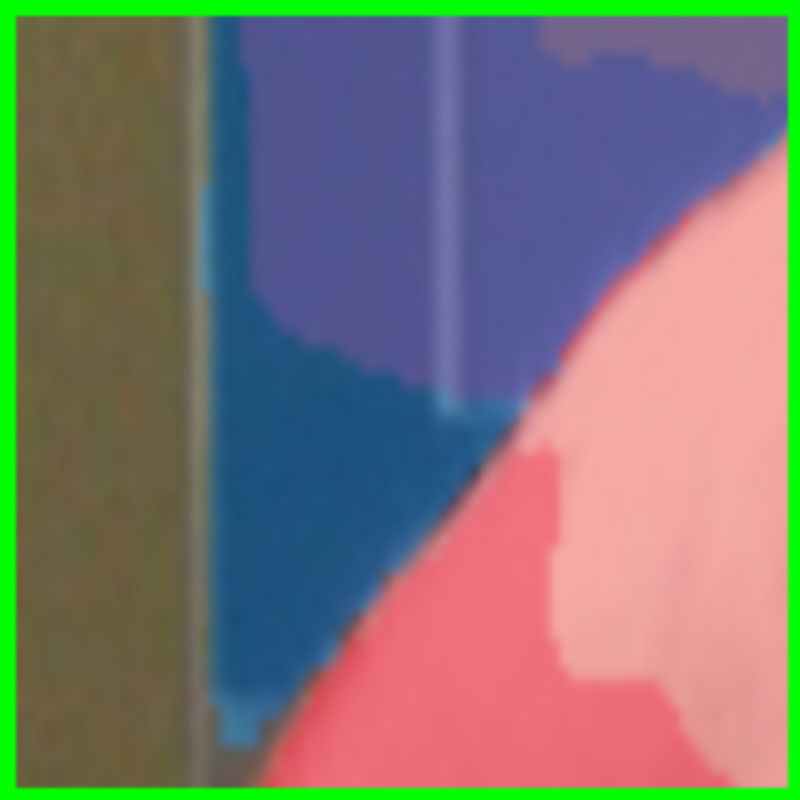}
        \end{subfigure}%
        \caption*{YOLACT}
      \end{subfigure}%
      \begin{subfigure}[b]{1.5cm}
        \begin{subfigure}[b]{1.5cm}
          \includegraphics[width=1.5cm]{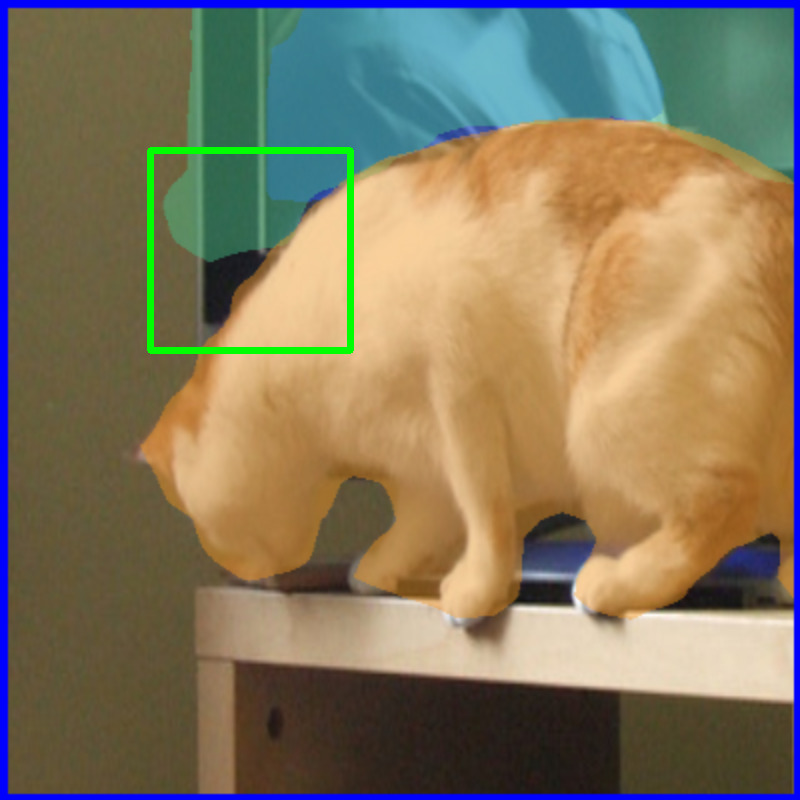}
        \end{subfigure}%
        \\[-0.2ex]
        \begin{subfigure}[b]{1.5cm}
          \includegraphics[width=1.5cm]{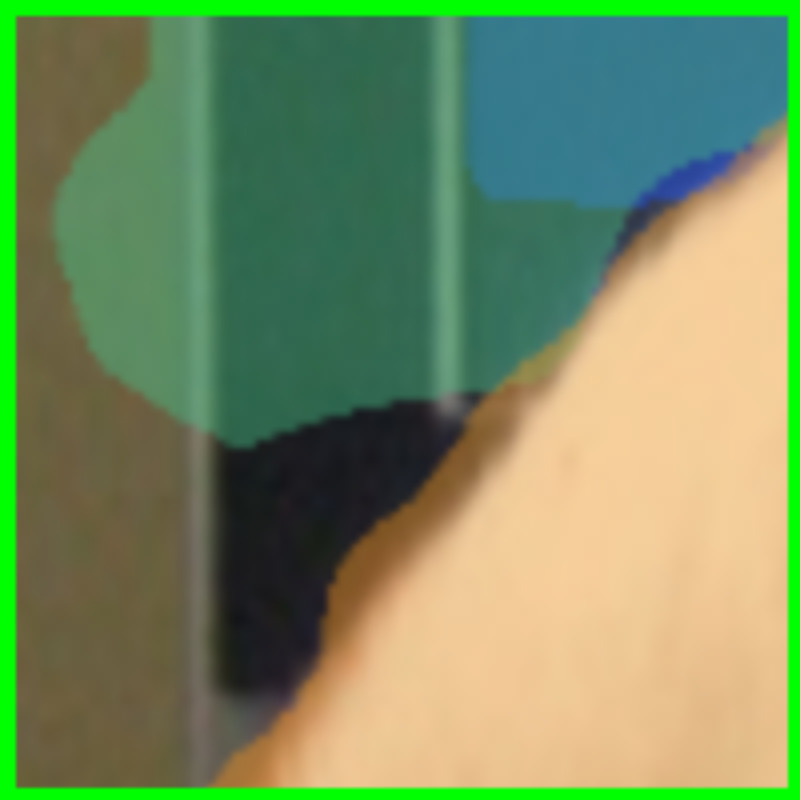}
        \end{subfigure}%
        \caption*{M-RCNN}
      \end{subfigure}%
  \end{subfigure}
  \begin{subfigure}[b]{7.5cm}
      \begin{subfigure}[b]{3cm}
        \includegraphics[width=3cm]{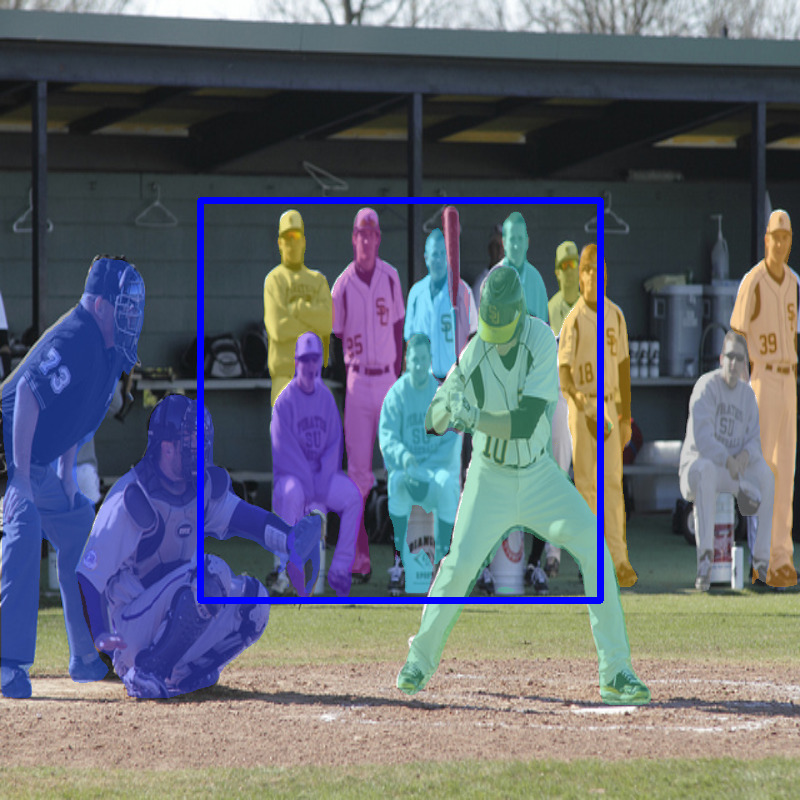}
        \caption*{ours}
      \end{subfigure}%
      \begin{subfigure}[b]{1.5cm}
        \begin{subfigure}[b]{1.5cm}
          \includegraphics[width=1.5cm]{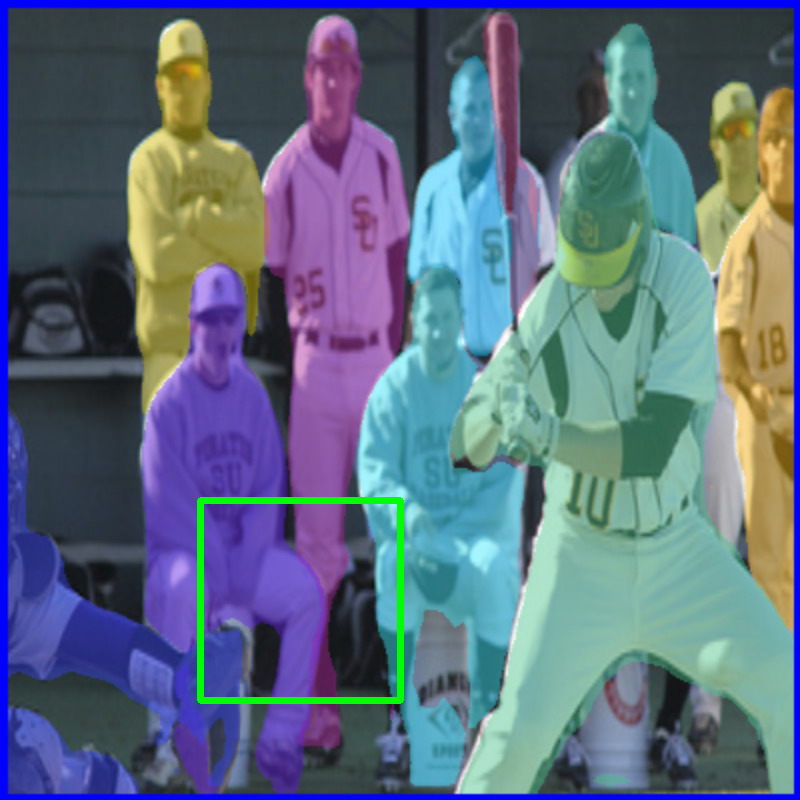}
        \end{subfigure}%
        \\[-0.2ex]
        \begin{subfigure}[b]{1.5cm}
          \includegraphics[width=1.5cm]{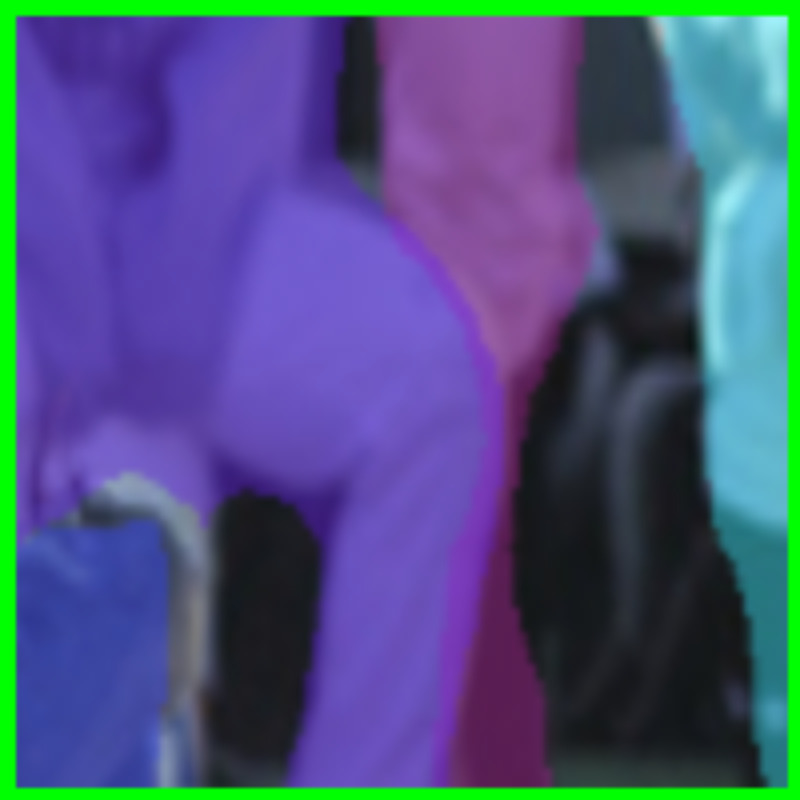}
        \end{subfigure}%
        \caption*{ours}
      \end{subfigure}%
      \begin{subfigure}[b]{1.5cm}
        \begin{subfigure}[b]{1.5cm}
          \includegraphics[width=1.5cm]{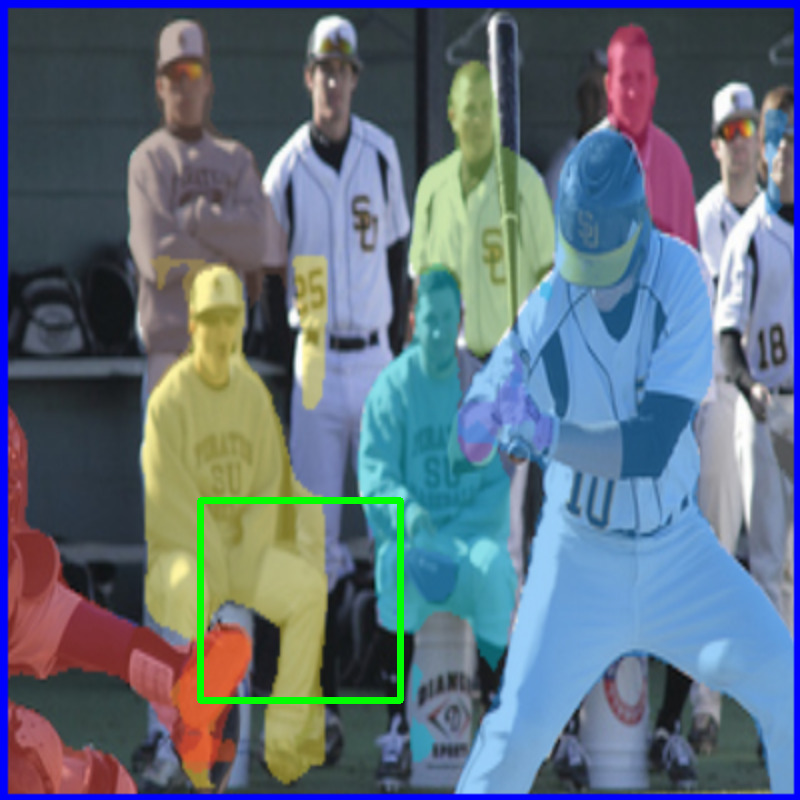}
        \end{subfigure}%
        \\[-0.2ex]
        \begin{subfigure}[b]{1.5cm}
          \includegraphics[width=1.5cm]{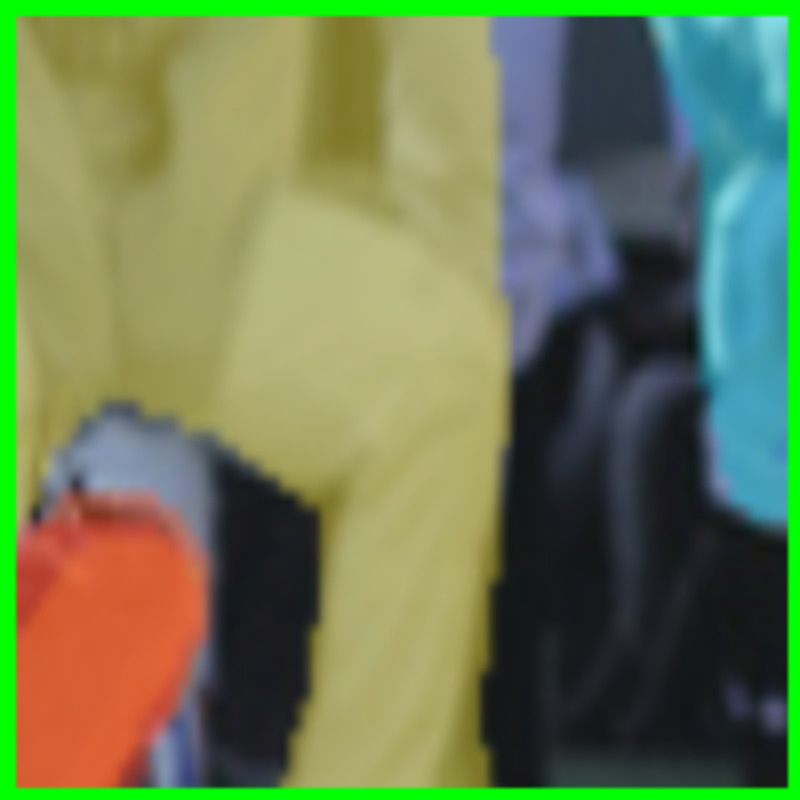}
        \end{subfigure}%
        \caption*{YOLACT}
      \end{subfigure}%
      \begin{subfigure}[b]{1.5cm}
        \begin{subfigure}[b]{1.5cm}
          \includegraphics[width=1.5cm]{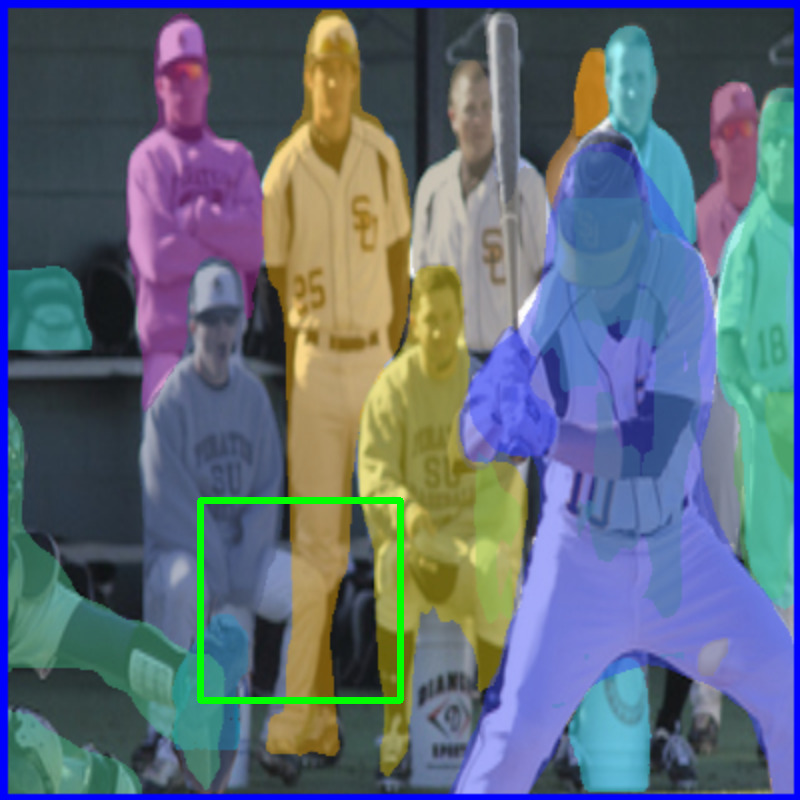}
        \end{subfigure}%
        \\[-0.2ex]
        \begin{subfigure}[b]{1.5cm}
          \includegraphics[width=1.5cm]{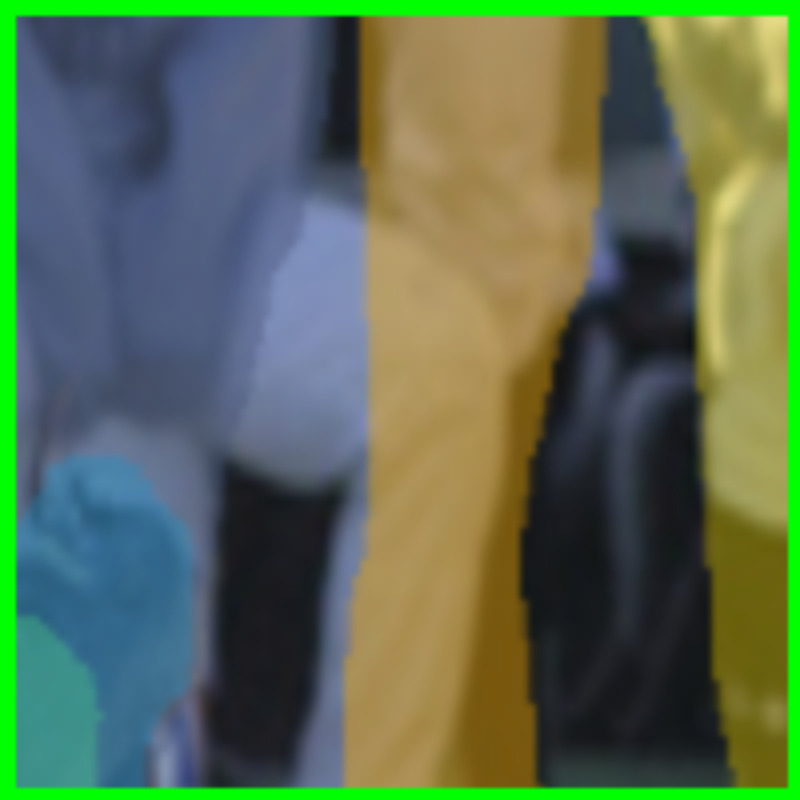}
        \end{subfigure}%
        \caption*{M-RCNN}
      \end{subfigure}%
  \end{subfigure}
} %
\caption{\textbf{Detailed comparison with other methods.} The large image on the left side is the segmentation result of our method. We further zoom in our result and compare
against
YOLACT~\cite{bolya2019yolact} (31.2\%  mAP) and Mask R-CNN~\cite{he2017mask} (36.1\%  mAP) on the right side. Our masks are 
overall 
of
higher quality.}
\label{fig:qualitative}
\end{figure*}

\subsection{Main result}\label{sec:qualitative}

\textbf{Quantitative results} We compare \OurMethod with Mask R-CNN~\cite{he2017mask} and TensorMask~\cite{chen2019tensormask} on the COCO \code{test}-\code{dev} dataset\footnote{To make fair comparison with TensorMask, the code base that we use for main result is \code{maskrcnn\_benchmark}. Recently released \code{Detectron2} fixed several issues of \code{maskrcnn\_benchmark} (\code{ROIAlign} and \code{paste\_mask}) in the previous repository and the performance is further improved.}. We use \code{56\_4\_14} with bilinear top interpolation, the DeepLabV3+ decoder with channel width 256 and P3, P5 input. Since our ablation models are heavily under-fitted, we increase the training iterations to 270K ($3 \times$ schedule), tuning learning rate down at 180K and 240K. Following Chen \etal's strategy~\cite{chen2019tensormask}, we use multi-scale training with shorter side randomly sampled from $[640, 800]$. As shown in Table~\ref{table:main}, our \OurMethod outperforms both the modified Mask R-CNN with deeper FPN and TensorMask using only half of their training iterations.

\OurMethod is also more efficient. Measured on a V100 GPU, the best R-101 \OurMethod runs at 0.07s/im, \textit{vs.} TensorMask's 0.38s/im, \textit{vs.} Mask R-CNN's 0.09s/im~\cite{chen2019tensormask}. Furthermore, a typical running time of our blender module is merely 0.6ms, which makes the additional time for complex scenes nearly
negligible
On the contrary, for two-stage Mask R-CNN with more expensive head computation, the inference time increases by a lot if the number of predicted instances grows.

\textbf{Real-time setting} We design a compact version of our model, \OurMethod-RT, to compare with YOLACT~\cite{bolya2019yolact}, a real-time instance segmentation method: i) the number of convolution layers in the prediction head is reduced to three, ii) and we merge the classification tower and box tower into one by sharing their features. We use Proto-FPN with four convolution layers with width 128 as the bottom module. The top FPN output P7 is removed because it has little effect on the detecting smaller objects. We train both \OurMethod-RT and Mask R-CNN with the $\times3$ schedule, with shorter side randomly sampled from $[440, 550]$.

There are still two differences in the implementation comparing to YOLACT. YOLACT resizes all images to square, changing the aspect ratios of inputs. Also, a paralleled NMS algorithm called Fast NMS is used in YOLACT. We do not adopt these two configurations because they are not conventionally used in instance segmentation researches. In YOLACT, a speedup of 12ms is reported by using Fast NMS. We instead use the Batched NMS in \code{Detectron2}, which could be slower than Fast NMS but does not sacrifice the accuracy. Results in Table~\ref{table:real-time} shows that \textit{\OurMethod-RT is 7ms faster and 3.3 AP higher than YOLACT-700}. Making our model also competitive under the real-time settings.

\textbf{Qualitative results} We compare our model with the best available official YOLACT and Mask R-CNN models with ResNet-101 backbone. Masks are illustrated in Figure~\ref{fig:qualitative}. Our model yields higher quality masks than Mask R-CNN. The first reason is that we predicts $56\times 56$ masks while Mask R-CNN uses $28\times 28$ masks. Also our segmentation module mostly utilizes high resolution features that preserve the original aspect-ratio, where Mask R-CNN also uses $28\times 28$ features.

Note  that YOLACT has 
difficulties 
discriminating instances of the same class close to each other. \OurMethod can avoid this typical leakage. This is because its top module provides more detailed instance-level information, guiding the bases to capture position-sensitive information and suppressing the outside regions.

\subsection{Discussions}\label{sec:mask rcnn}
\textbf{Comparison with Mask R-CNN} Similar to Mask R-CNN, we use RoIPooler to locate instances and extract features. We reduce the running time by moving the computation of R-CNN heads before the RoI sampling to generate position-sensitive feature maps. Repeated mask representation and computation for overlapping proposals are avoided. We further simplify the global map representation by replacing the hard alignment in R-FCN~\cite{dai2016r} and FCIS~\cite{li2017fully} with our attention guided blender, which needs ten times less channels for the same resolution.

Another advantage of \OurMethod is that it can produce higher quality masks, since our output resolution is not restricted by the top-level sampling. Increasing the RoIPooler resolution of Mask R-CNN will introduce the following problem. The head computation increases quadratically with respect to the RoI size. Larger RoIs requires deeper head structures. Different from dense pixel predictions, RoI foreground predictor has to be aware of whole instance-level information to distinguish foreground from other overlapping instances. Thus, the larger the feature sizes are, the deeper sub-networks is needed.

Furthermore, it is not very friendly to real-time applications that the inference time of Mask R-CNN is proportional to the number of detections. By contrast, our blender module is very efficient (0.6ms on 1080 Ti). The additional inference time required after increasing the number of detections can be neglected.

Our blender module is very flexible. Because our top-level instance attention prediction is just a single convolution layer, it can be an almost free add-on to most modern object detectors.  With its accurate instance prediction, it can also be used to refine two-stage instance predictions.

\subsection{Panoptic Segmentation}
\label{sec:Panoptic}

We use the semantic segmentation branch of Panoptic-FPN~\cite{kirillov2019panoptic} to extend \OurMethod to the panoptic segmentation task. We use annotations of COCO 2018 panoptic segmentaiton task. All models are trained on \code{train2017} subset and tested on \code{val2017}. We train our model with the default FCOS~\cite{tian2019fcos} $3\times$ schedule with scale jitter (shorter image side in $[640, 800]$. To combine instance and semantic results, we use the same strategy as in Panoptic-FPN, with instance confidence threshhold 0.2 and overlap threshhold 0.4.

Results are reported in Table~\ref{table:panoptic}. Our model is consistently better than its Mask R-CNN counterpart, Panoptic-FPN. We assume there are three reasons. First, our instance segmentation is more accurate, this helps with both thing and stuff panoptic quality because instance masks are overlaid on top of semantic masks. Second, our pixel-level instance prediction is also generated from a global feature map, which has the same scale as the semantic prediction, thus the two results are more consistent. Last but not least, since the our bottom module shares structure with the semantic segmentation branch, it is easier for the network to share features during the closely related multi-task learning.

\begin{table*}[t!]
\centering
\small
\begin{tabular}{r |c|ccc|ccccc}
\hline
Method &Backbone & PQ & SQ & RQ & PQ\textsuperscript{Th} & PQ\textsuperscript{St} & mIoU & AP\textsuperscript{box} & AP\\
\hline
\hline
Panoptic-FPN~\cite{kirillov2019panoptic} &\multirow{2}{*}{R-50} & 41.5 & 79.1 & 50.5 & 48.3 & 31.2 & 42.9 & 40.0 & 36.5\\
BlendMask & & \textbf{42.5} & \textbf{80.1} & \textbf{51.6} & \textbf{49.5} & \textbf{32.0} & \textbf{43.5} & \textbf{41.8} & \textbf{37.2}\\
\hline
Panoptic-FPN \cite{kirillov2019panoptic} &\multirow{2}{*}{R-101}& 43.0 & 80.0 & 52.1 & 49.7 & 32.9 & 44.5 & 42.4 & 38.5\\
BlendMask & & \textbf{44.3} & \textbf{80.1} & \textbf{53.4} & \textbf{51.6} & \textbf{33.2} & \textbf{44.9} & \textbf{44.0} & \textbf{38.9}\\
\hline
\end{tabular}
\caption{\textbf{Panoptic results} on COCO \code{val2017}. Panoptic-FPN results are from the official \code{Detectron2} implementation,
which are improved upon the original published results in \cite{kirillov2019panoptic}. 
}
\label{table:panoptic}
\end{table*}

\subsection{More Qualitative Results}
\begin{figure*}[ht]
\centering
\includegraphics[width=.985\textwidth]{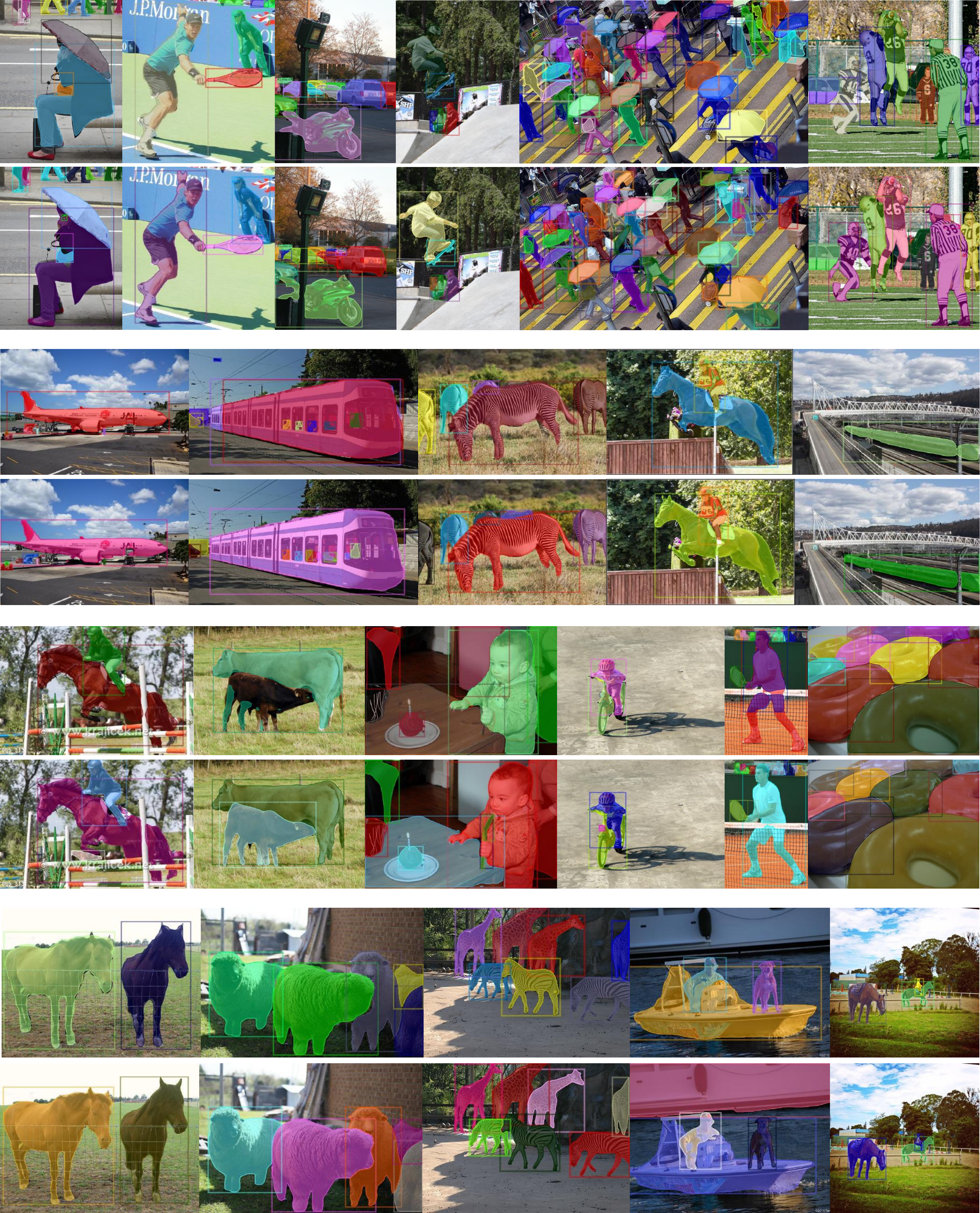}
\caption{Selected results of Mask R-CNN (top) and \OurMethod (bottom). Both models are based on \code{Detectron2}. The Mask R-CNN model is the official 3$\times$  R101 model with 38.6 AP. \OurMethod model obtains 39.5 AP. Best viewed in digital format with zoom.}
\label{fig:coco-res}
\end{figure*}

We visualize qualitative results of Mask R-CNN and \OurMethod on the validation set in Fig.~\ref{fig:coco-res}. Four sets of images are listed in rows. Within each set, the top row is the Mask R-CNN results and the bottom is \OurMethod. Both models are based on the newly released \code{Detectron2} with use R101-FPN backbone. Both are trained with the $3\times$ schedule. The Mask R-CNN model achieves 38.6\%  AP and ours 39.5\%  AP.

Since this version of Mask R-CNN is a very strong baseline, and both models achieve very high accuracy, it is very difficult to tell the differences. To demonstrate our advantage, we select some samples where Mask R-CNN has trouble dealing with. Those cases include:

\begin{itemize}
\itemsep -0.1cm

    \item Large objects with complex shapes (Horse ears, human poses). Mask R-CNN fails to provide sharp borders.
    \item Objects in separated parts (tennis players occluded by nets, trains divided by poles). Mask R-CNN tends to include occlusions as false positive or segment targets into separate objects.
    \item Overlapping objects (riders, crowds, drivers). Mask R-CNN gets uncertain on the borders and leaves larger false negative regions. Sometimes, it assigns parts to the wrong objects, such as the last example in the first row.
\end{itemize}

Our \OurMethod performs better on these cases. 1) Generally, \OurMethod utilizes features with higher resolution. Even for the large objects, we use stride-8 features. Thus details are better preserved. 2) As shown in previous illustrations, our bottom module acts as a class agnostic instance segmenter which is very sensitive to borders. 3) Sharing features with the bounding box regressor, our top module is very good at recognizing individual instances.
It can generate attentions with flexible shapes to merge the fine-grained segments of bottom module outputs.

\subsection{Evaluating on LVIS annotations}
To quantify the high quality masks generated by \OurMethod{}, we compare our results with on the higher-quality LVIS annotations~\cite{gupta2019lvis}. Our model is compared to the best high resolution model we are aware of, recent PointRend~\cite{kirillov2019pointrend}, which uses multiple subnets to refine the local features to get higher resolution mask predictions. The description of the evaluation metric can be found in~\cite{kirillov2019pointrend}. Table~\ref{table:lvis} shows that the evaluation numbers will improve further given more accurate ground truth annotations. Our method can benefits from the accurate bottom features and surpasses the high-res PointRend results.

\begin{table*}[t!]
\centering
\small
\begin{tabular}{r |c|c|cc}
\hline
Method &Backbone & resolution & COCO AP & LVIS AP$^\star $\\
\hline
\hline
Mask R-CNN & X101-FPN & $28\times28$ & 39.5 & 40.7\\ 
PointRend & X101-FPN & $224\times224$ & 40.9 & 43.4\\ 
\OurMethod{} & R-101+dcni3 & $56\times56$ & \textbf{41.1} & \textbf{44.1}\\ 
\hline
\end{tabular}
\caption{\textbf{Comparison with PointRend}. Mask R-CNN and PointRend results are quoted from Table 5 of ~\cite{kirillov2019pointrend}. Our model is the last model in Table~\ref{table:main}. Our model is 0.2 points higher on COCO and 0.7 points higher 
on LVIS annotations.
Here 
LVIS AP$^\star $ is 
COCO mask AP evaluated against
the higher-quality LVIS annotations.
}
\label{table:lvis}
\end{table*}

\section{Conclusion}

We have devised a novel blender module for instance-level dense prediction tasks which uses  both high-level instance and low-level semantic information. It is efficient and easy to integrate with different
main-stream detection networks.

Our framework \OurMethod outperforms the
care\-fully-en\-gi\-neered Mask R-CNN without bells and whistles while being $20\%$ faster.
Furthermore,
the real-time version \OurMethod-RT achieves 34.2\%  mAP at 25 FPS evaluated on a single 1080Ti GPU card. We believe that our \OurMethod  is capable of  serving  as an
alternative to Mask R-CNN~\cite{he2017mask}  for many other instance-level recognition tasks.

\section*{Acknowledgements}

The authors would like to thank
Huawei Technologies for the donation of
GPU cloud computing resources.

{\small
	\bibliographystyle{ieee_fullname}
	\bibliography{ref}
}

\end{document}